\pdfoutput=1

\documentclass[11pt]{article}

\usepackage[]{ACL2023}

\usepackage{times}
\usepackage{latexsym}

\usepackage[T1]{fontenc}

\usepackage[utf8]{inputenc}

\usepackage{microtype}

\usepackage{inconsolata}
\usepackage{times}
\usepackage{latexsym}
\usepackage{amsmath}
\usepackage{multirow}
\usepackage{subfigure}
\usepackage{color}
\usepackage[ruled,linesnumbered]{algorithm2e}
\usepackage{graphicx}
\usepackage{tabu}
\usepackage{booktabs}
\usepackage{amsfonts}
\usepackage{makecell}
\usepackage{pgf}
\usepackage{tikz}
\usepackage{pifont}
\usepackage{amsmath,amssymb}
\usetikzlibrary{positioning}

\title{Towards Graph-hop Retrieval and Reasoning in \\ Complex Question Answering over Textual Database} 

\author{Minjun Zhu$^{1,2}$\thanks{These authors contribute this work equally.}, Yixuan Weng$^{*1}$, Shizhu He$^{1,2}$, Kang Liu$^{1,2}$, Jun Zhao$^{1,2}$ \\
	$^1$ National Laboratory of Pattern Recognition,, Institute of Automation, CAS \\
	$^2$ School of Artificial Intelligence, University of Chinese Academy of Sciences\\
\texttt{zhuminjun2020@ia.ac.cn,wengsyx@gmail.com, \{shizhu.he, kliu, jzhao\}@nlpr.ia.ac.cn}
}

\begin{document}
\maketitle
\begin{abstract} 
 

In Textual question answering (TQA) systems, complex questions often require retrieving multiple textual fact chains with multiple reasoning steps. While existing benchmarks are limited to single-chain or single-hop retrieval scenarios. In this paper, we propose to conduct \textbf{Graph-Hop} \textemdash a novel multi-chains and multi-hops retrieval and reasoning paradigm in complex question answering. We construct a new benchmark called \textbf{ReasonGraphQA}, which provides explicit and fine-grained evidence graphs for complex questions to support interpretable reasoning, comprehensive and detailed reasoning. And ReasonGraphQA also shows an advantage in reasoning diversity and scale. Moreover, We propose a strong graph-hop baseline called \textbf{B}idirectional \textbf{G}raph \textbf{R}etrieval (\textbf{BGR}) method for generating an explanation graph of textual evidence in knowledge reasoning and question answering. We have thoroughly evaluated existing evidence retrieval and reasoning models on the ReasonGraphQA. Experiments highlight Graph-Hop is a promising direction for answering complex questions, but it still has certain limitations. We have further studied mitigation strategies to meet these challenges and discuss future directions.\footnote{Dataset and codes will be made publicly available upon acceptance.}


\end{abstract}

\section{Introduction}

Retrieving and reasoning about world knowledge is the core ability of question answering (QA) task \cite{NitishGupta2019NeuralMN}. Textual question answering (TQA) systems need to retrieve relevant evidence and conduct knowledge reasoning \cite{DanqiChen2017ReadingWT} on answering complex questions over multiple passages or facts \cite{ZhilinYang2018HotpotQAAD,qi-etal-2021-answering}. Recently, lots of tasks and datasets have been proposed and sparked significant progress of TQA in different scenarios \cite{FengbinZhu2021RetrievingAR,ZhilinYang2018HotpotQAAD,JamesThorne2021DatabaseRO}. 

\begin{figure}[t]
	\centering
	\includegraphics[scale=0.32]{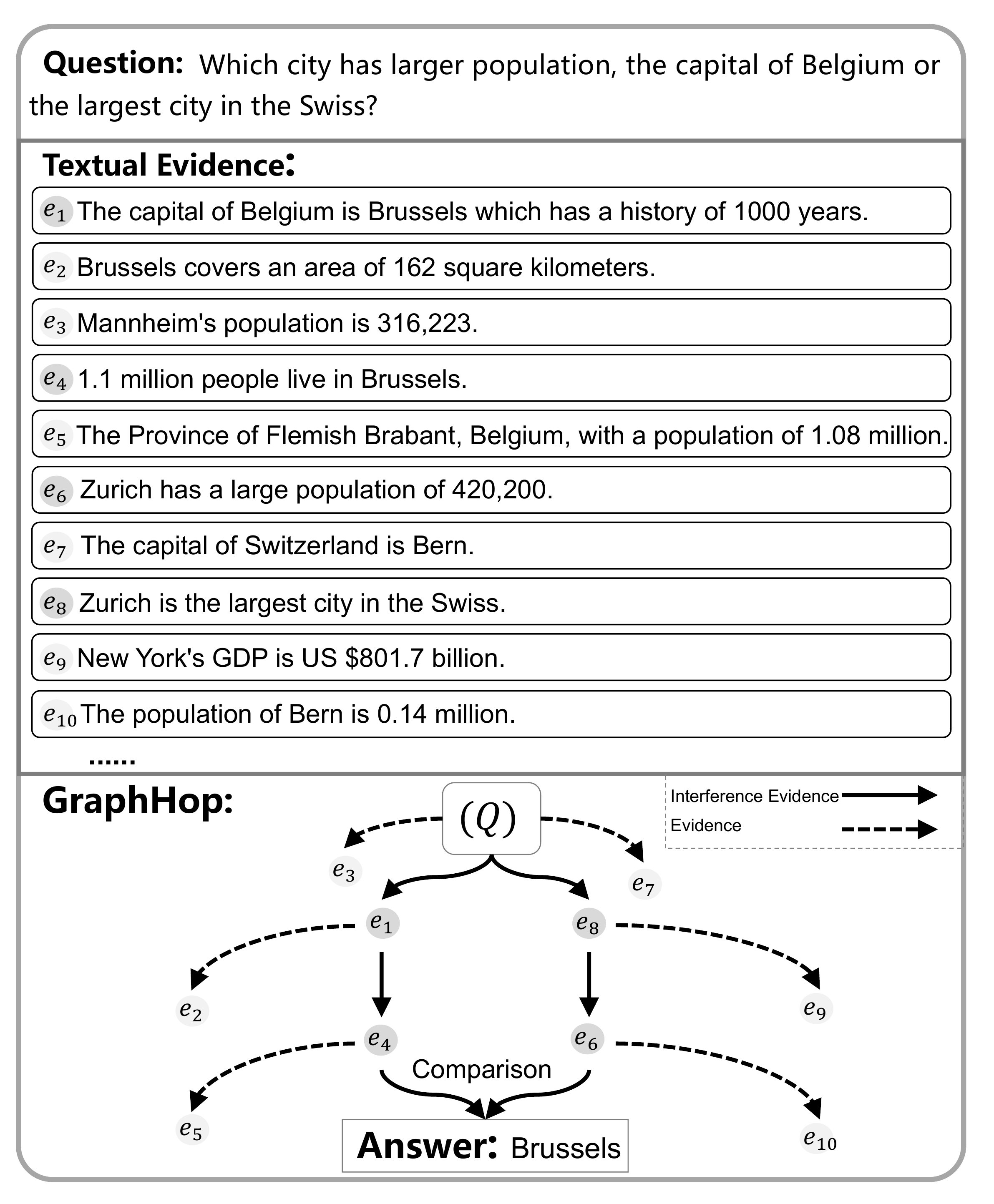}
	\caption{An example of ReasonGraphQA, it requires multiple chains of fact sets and each chain involves two-hop reasoning in answering this complex question.} 
	\label{f1}
	\vspace{-0.5cm}
\end{figure}
\begin{table*}[t]
		\centering 
  
		\renewcommand\arraystretch{1.2}
  \resizebox{0.8\textwidth}{!}{%
		\begin{tabular}{l|cccccc}
			\noalign{\hrule height 1pt}

			\multicolumn{1}{c|}{Dataset}  & Reasoning Types& Evidence &Text Type& Multi-Chains& Multi-Hops & Evidence Structures\\
						\hline
                TriviaQA \cite{joshi-etal-2017-triviaqa} & -& \ding{55} &passage& \ding{55}& \ding{55} & 1\\
			HotPotQA \cite{ZhilinYang2018HotpotQAAD} &3&$\checkmark$&passage&\ding{55}& \ding{55} & 1   \\
               BeerQA \cite{qi-etal-2021-answering} &3&\ding{55}&passage&\ding{55}& $\checkmark$ &3+ \\
                WikiNLDB \cite{JamesThorne2021DatabaseRO}&4&$\checkmark$&sentence&\checkmark& \ding{55} & 2\\
                eQASC \cite{jhamtani-clark-2020-learning}&2&$\checkmark$&sentence&\ding{55}& $\checkmark$&1\\
                ReasonGraphQA (Ours)&\textbf{5}&$\checkmark$&sentence&$\checkmark$&$\checkmark$&\textbf{262}\\
			\noalign{\hrule height 1pt}
		\end{tabular}}
		\caption{Comparison ReasonGraphQA with existing datasets of Complex TQA.}
		\vspace{-0.4cm}
		\label{data_compare}
	\end{table*}

However, those datasets still have some limitations. On the one hand, most open domain question answering (ODQA) only focus on multi-hop reasoning of a single chain. For example, HotpotQA \cite{ZhilinYang2018HotpotQAAD} devotes to addressing two-hop questions and BeerQA \cite{qi-etal-2021-answering} requires a varying number of retrieval steps over multiple passages. The above only focus on one-chain retrieval and reasoning. On the other hand, some textual datasets include multiple discretization chains but only requires single-hop reasoning to answer question, such as WIKINLDB \cite{JamesThorne2021DatabaseRO} and eQASC \cite{jhamtani-clark-2020-learning}. Moreover, compared with knowledge-based question answering (KBQA), current TQA datasets do not thoroughly test the complexity and diversity of question types and reasoning types \cite{KQA}.



In fact, answering complex questions often requires a combination of retrieving multi-chains and using multi-hops reasoning to infer the answer. We refer to this process as Graph-Hop retrieval and reasoning (shorted as \textbf{Graph-Hop}). As shown in figure \ref{f1}, to answer this question, system first retrieves the population of each city (multi-chain), and then uses multi-hop reasoning on each chain to infer the population value. Finally, it compares two values and identifies the city with the larger population. This process requires an evidence graph with multiple chains and hops. In this way, Graph-Hop provides a more fine-grained and adaptable representation for complex question answering tasks.


In addition, existing textual question answering systems still have trouble explaining explicitly why an answer is correct or not and ``how'' the answer is obtained step-by-step. Although the existing retrieval methods can directly retrieve the relevant passages \cite{XiangyangMou2021ComplementaryEI,KoustavRudra2021AnIA,lu-etal-2021-multi,zhu2022reasonchainqa}, they cannot retrieve a structured evidence graph, which limits the ability of the model's reasoning and interpretation.

To address above problems, we introduce a benchmark called ReasonGraphQA and provides explanation evidence graphs to explicitly describe the reasoning process for solving complex questions. Evidence graphs can provide intermediate results and facilitate human understanding. It also allows for better control of the model behavior, enabling users to easily identify errors by inspecting the outputs of intermediate steps. Moreover, compared with other datasets (as shown in Table \ref{data_compare}), ReasonGraphQA not only contains more diversified hops and chains evidences, but also cover more reasoning types of complex questions and richer explicitly evidence structures.

We also propose a specific \textbf{B}idirectional \textbf{G}raph \textbf{R}etrieval (\textbf{BGR}) method to support Graph-Hop. This method retrieves evidence from both forward and backward directions, and then fuses them to construct evidence graphs and support to answer complex questions. We compared four types of retrieval and reasoning systems on the ReasonGraphQA dataset. Experimental results have shown that BGR achieved strong performance in both the retrieval task and the explanation graph task. However, their performance is still far from human-level performance in the explanation graph construction task, it is suggesting that further research should consider more on Graph-Hop.

In summary, our contributions are as follows: \textbf{(1)} We propose a Graph-Hop paradigm and construct a new benchmark ReasonGraphQA, which includes diverse question types and explicit reasoning processes to guide interpretable retrieval and question answering over textual databases in a fine-grained and comprehensive way. \textbf{(2)} We also propose a Bidirectional Graph Retrieval (\textbf{BGR}) method, which utilizes both forward reasoning and backward reasoning information. \textbf{(3)} Our evaluation of four retrieval systems on ReasonGraphQA demonstrates that Graph-Hop Retrieval is a promising approach. We also discuss potential future directions to address Graph-Hop challenges.

\begin{figure*}[t]
	\centering
	\includegraphics[scale=0.578]{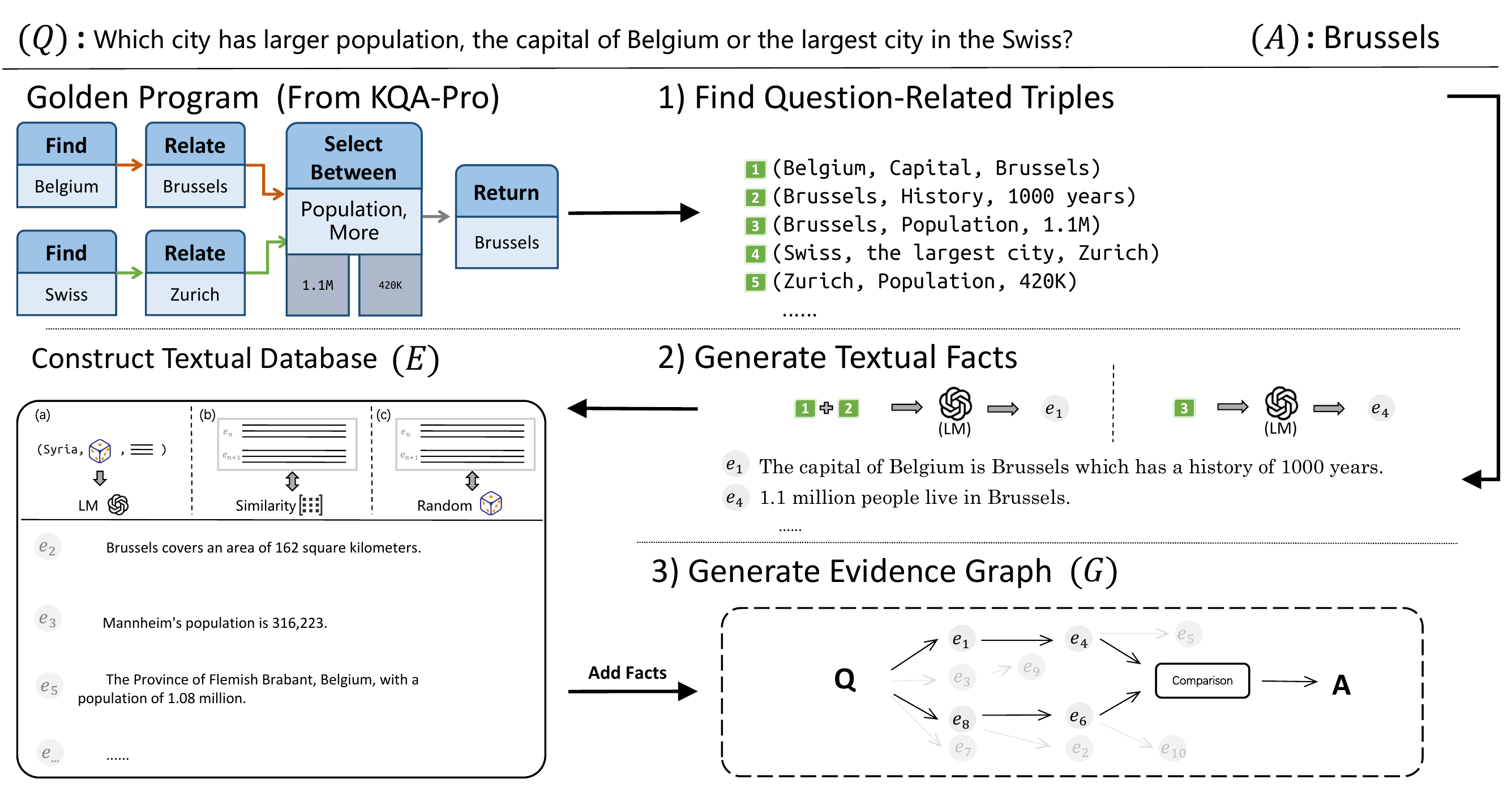}
    \caption{ReasonGraphQA construction process. We use Golden Program to generate explanation  evidence graphs and create a text database for each question-answer pair. It consisting of three steps: finding question-relevant triples, generating textual evidence, and generating an explanation  evidence graph.}
	\label{dataset_process}
	\vspace{-0.34cm}
\end{figure*}

\section{Related Work}
Textual question answering (TQA) requires to retrieve evidence from a large corpus to answer natural language questions. Some researchers proposed a novel TQA task over natural language database (NLDB) and support natural language database queries such as filtering, comparison, and aggregation, where database consists of unordered sets of textual facts \cite{JamesThorne2021DatabaseRO,zhu2022reasonchainqa}. It requires comprehensive reasoning and retrieval of text sentences  \cite{TomerWolfson2020BreakID}. Despite the rapid progress in TQA, they ignore the problem of multi-hop retrieval in multi-chain fact sets that may appear in complex textual question answering. In comparison, the proposed ReasonGraphQA requires graph retrieval from large-scale textual databases. And we focus on discrete reasoning over textual evidence, which greatly evaluate the structured path modeling and discrete reasoning ability of QA systems over the textual databases. (More comprehensive related work is shown in Appendix \ref{comprehensivework})

\section{Graph-Hop Over Textual Database}
ReasonGraphQA devotes to answering complex questions that need Graph-Hop (multi-hop multi-chain) over database. Both question and evidences of database are represented as natural language sentences, each sentence is stand-alone and contains one or multiple facts. Formally, given a question $Q$ and a textual database $E = \{e_1, \dots, e_n\}$, system needs to: (1) retrieve an explicable reasoning graph $G$ from the given textual database, (2) obtain the answer $A$ based on the explanation  graph $G$; The graph $G$ is a directed acyclic graph composed of the evidences in $E$ that are related to the question and used to reason the answer. 

\section{Construction of ReasonGraphQA} 
In this section, we present the construction process of ReasonGraphQA dataset. Constructing fine-grained evidence graphs for complex questions is a non-trivial task. We develop an approach to automatically construct a dataset with complex questions, answers and explanation evidence graphs . 
Figure \ref{dataset_process} illustrates the main construction process of ReasonGraphQA using the example in Figure \ref{f1}.

\subsection{Question-related Triples Finding\label{Finding Triples}}

We obtain complex questions and answers from the KQA-Pro dataset \cite{KQA}, a large-scale KBQA dataset \cite{PeiyunWu2019ASO}, which requires reasoning over multiple pieces of evidence. To automate the generation of question-related evidence, we use structured queries ``KoPL program'' of the KQA-pro dataset and ground each programming procedure to Wikidata triples. As illustrated in Fig. \ref{dataset_process} (1), the structured Golden Program, consisting of ``Relate'', ``Find'', and ``Select between'' operations, can identify five triples of Wikidata. By searching the target knowledge base (e.g., Wikidata), we can obtain factual facts needed to answer the question.

\subsection{Textual Facts Generation \label{Evidence Generation}}

Based on data-to-text work~\cite{agarwal-etal-2021-knowledge}, we can convert the structured facts into unstructured texts. To improve the diversity, naturalness, and information of the generated text, we propose a method of building triple subgraphs by selecting 0-2 triples with the same head entity from Wikidata according to a certain probability and combine them into a subgraph. While ensuring that they do not overlap with other subgraphs to make sure textual facts remain independent. The subgraphs are then input into a pre-trained language model (T5) fine-tuned on the KELM\cite{agarwal-etal-2021-knowledge} corpus to generate unstructured text. As shown in Figure \ref{dataset_process} (2). To ensure completeness of entities in the triples, we use string matching to exclude missing text, and use BERTScore \cite {TianyiZhang2019BERTScoreET} to select the most appropriate text evidence from multiple generated options as the correct evidence.


\subsection{Textual Database Construction \label{Database Construction}}
We obtain a large-scale textual database containing generated evidences (\ref{Evidence Generation}). For each question, we can retrieve evidence from those large-scale sentences (e.g., more than 100 billion sentences). However, in our experimental environment (500000 sentences in total), we must consider computing efficiency and retrieval cost. Therefore, we have retrieved an appropriate number of sentences from the complete textual database to form a target textual database from which we select evidence for each question. Specifically, apart from the golden evidence, we also retrieve other sentences that are related to the question to form the target textual database. Additionally, to construct a task closer to the real retrieval scene, and to verify knowledge-based reasoning ability, we have added interference evidence to the database. In this paper, the interference-related evidence is obtained from the following three categories of methods (1/3 of each category): (a) SimCSE \cite{gao-etal-2021-simcse} is used to select evidence with similar semantics of the question; (b) We use the same head entity but different relation triples to regenerate evidence sentences; (c) We randomly select other textual evidence.

\subsection{Evidence Graph Generation \label{Reason Graph Generation}} 
The reasoning graph of textual evidence is the key component of ReasonGraphQA. We extract and re-summarize the structure among golden triples with the programming language ``KoPL program'', and utilize network  \footnote{https://networkx.org} to build the reasoning graph of sentences. In order to ensure the high quality of the evidence graph, we carefully follow these constraints during its construction. (1) Each evidence contains at least one knowledge fact; (2) Each question must be answered with a clear reasoning explanation  graph $G$; (3) Each graph $G$ must be a directed acyclic graph; (4) Any non-leaf node has at least one path to the root node; (5) All evidence cannot be repeated on the path to the root node (avoiding loops). Samples that do not meet these constraints are removed. An example of evidence graph is shown in figure \ref{dataset_process}. (3), which reflects the reasoning progress from question to answer.

\subsection{Dataset Analysis}
\begin{table}[t]
		\centering \small
		\renewcommand\arraystretch{1.1} \setlength{\tabcolsep}{1.4mm}
		\begin{tabular}{c|cccc|c}
			\noalign{\hrule height 1pt}

			\textbf{Dataset}  & SC,SH& SC,MH&MC,SH& MC,MH& Number \\
						\hline
			\textbf{Train} &2,295&3,524&1,001&3,883& 11,703  \\
			\textbf{Dev} &321&577&141&467& 1,506 \\
			\textbf{Test}&248&572&135&514 & 1,469  \\
			\hline
			\textbf{Total}  &2,864&5,673&1,277&4,864&14,678 \\
			\noalign{\hrule height 1pt}
			
		\end{tabular}
		\caption{The statistics of ReasonGraphQA, where SC, MC, SH and MH indicate single-chain, multi-chain, single-hop, and multi-hop, respectively. }
		\vspace{-0.4cm}
		\label{data}
	\end{table}
The ReasonGraphQA dataset consists of 14,678 examples, which are divided into training (11,703), dev (1,506), and test (1,469) sets using a random probability of 8:1:1. Table \ref{data} presents statistics on the graph size and structure of the dataset. The dataset includes four types of evidence graphs: ``single-chain single-hop,'' ``single-chain multi-hop,'' ``multi-chain single-hop,'' and ``multi-chain multi-hop,'' which account for 19.5\%, 38.6\%, 8.7\%, and 33.2\% of the dataset, respectively. There are 262 nonisomorphic graph structures in the dataset. The questions in the dataset are classified into five types: ``query'', ``comparison'', ``count'', ``boolean'', and ``qualifier'' based on nine asking strategies used in original KQA-Pro dataset (details in Appendix~\ref{sec:appendix_dataanalysis}). These diverse graph structures provide more detailed and interpretable evidence for complex questions.

\subsection{Quality Evaluation}
To evaluate the quality of mapping facts from knowledge triples, 500 sampled facts were scored based on smoothness, faithfulness, and sufficiency. 98.2$\%$ (491/500) facts were smooth, with only 9 containing repeated text. 98.6$\%$ (493/500) facts were faithful to the relation of the triples, with only 7 containing additional information. Three facts replaced incorrect information with correct information, resulting in a faithfulness and sufficiency score of 0. The remaining four facts contained additional information that enriched the context. 

We conducted manual evaluation and found that the quality of the data set construction is relatively high. For example, 96$\%$ of facts in WiKiNLDB are loyal to relationships, while ReasonGraphQA is 98.6$\%$. This demonstrates that the data set presented in this paper is suitable for model development and technical verification of complex question answering in textual databases (details in Appendix~\ref{sec:appendix_dataanalysis}). 



\section{Methods}
\begin{figure}[h]
	\centering
	\includegraphics[scale=0.307]{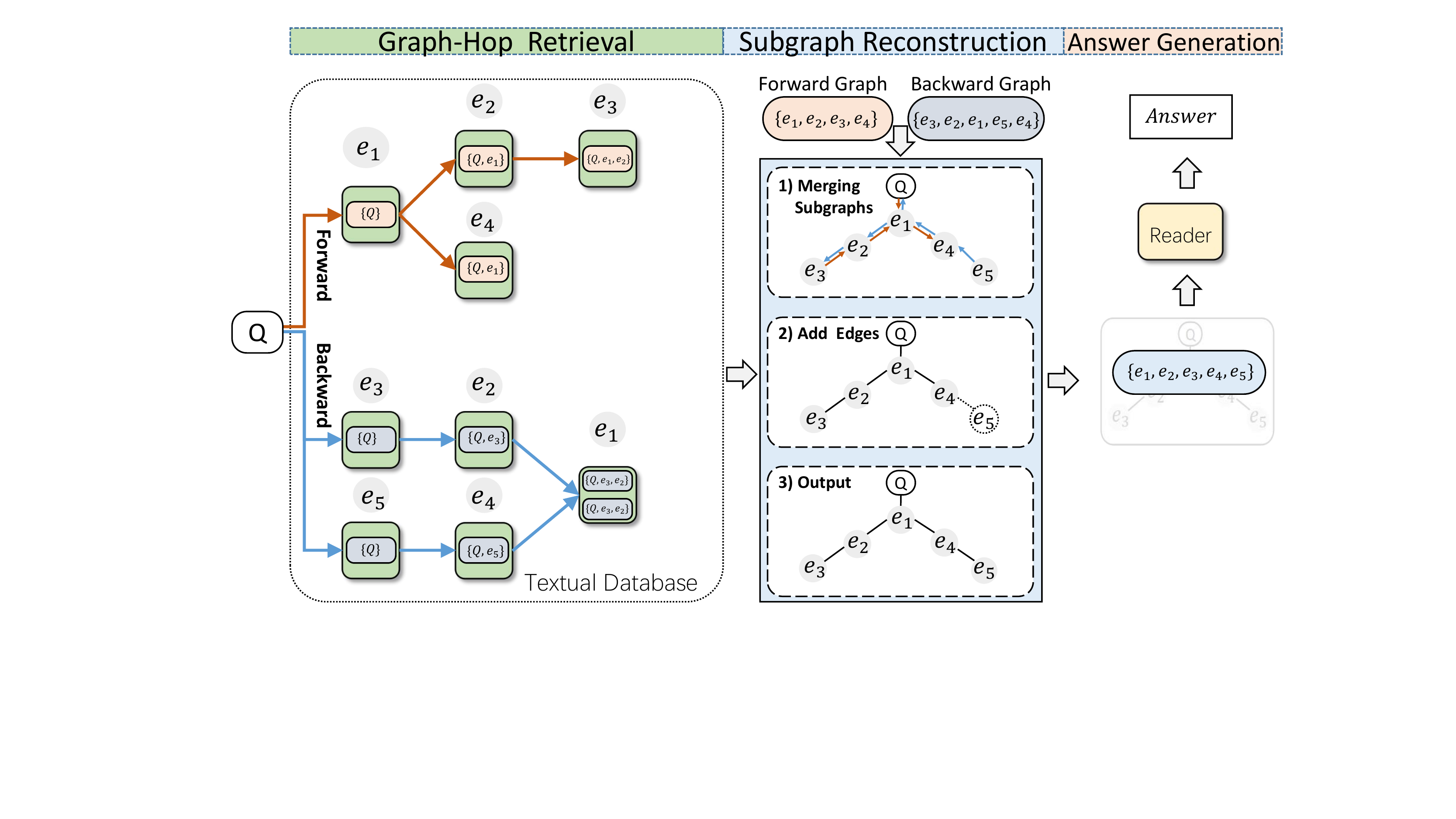}
    \caption{Overview of our proposed Bidirectional Graph-hop Retrieval (BGR) method.}
	\label{23}
	\vspace{-0.34cm}
\end{figure}

In this section, we present our proposed retrieval-based question-answering model. This model follows the popular retrieval-reader architecture. Figure \ref{23} illustrates the architecture of our model, which consists of Bidirectional Graph-Hop Retrieval, Subgraph Reconstruction, and Answer Generation.

\subsection{Bidirectional Retrieval}
We design a bidirectional retrieval method to improve graph-hop retrieval accuracy. "In traditional chain retrieval, the model starts by searching for the first relevant evidence, and continues iteratively. However, the structure of evidence in graph retrieval is more complex,  resulting in a higher error rate as the search depth increases. To mitigate this issue, we introduce backward retrieval, which starts at the leaf nodes of evidence graph and searches evidence from back to front. As a result, we obtain two evidence subgraphs,one from forward retrieval and one from backward retrieval, as depicted in Figure \ref{23}. By merging these two subgraphs, our bidirectional retrieval method can mitigate the problem of rapidly declining accuracy in the forward retrieval with increasing depth in the graph.

Given a question $Q$ and a candidate evidence base $E=\{e_1, e_2,\cdots, e_n\}$, we represent $Q$ and $E$ using BERT to obtain their representations, $h_0 = \text{BERT}(Q)$. The retrieval process follows a depth-first search, where at each step, the current evidence node $i$ may have multiple paths that are reachable. These paths are represented as $H_i=\{h_i^1,\cdots,h_i^{i_k}\}$, where $i_k$ is the number of paths per node. These paths are matched one by one with the path code and evidence base $E_i$ ($E_i \subseteq  E$). To handle the complex structure of graph retrieval, we use a feedforward neural network (composed of linear layers and activation functions) instead of a similarity threshold to match the next layer of evidence nodes. Every time a new evidence node is retrieved, we use the Attention mechanism to combine the path set $H_i$ and the retrieved evidence node $e_{i+1}$ to generate a new path set $H_{i+1}$. The whole process is illustrated in Figure \ref{23}. We repeat this process until no new nodes can be retrieved.

{\small \begin{eqnarray}
{ \widetilde{E}_i=\cup E L U\left(F F N\left(h_i, E_i\right)\right) \quad h_i \in H_i} \end{eqnarray}}
\vspace{-0.4cm}
{\small \begin{eqnarray}
{\widetilde{E}_{i+1}=E_{i+1} \cup \widetilde{E}_i, \quad H_{i+1}=\left\{h_i \cup \tilde{e}_i \mid \tilde{e}_i \in \tilde{E}_i\right\} }
\end{eqnarray}}

\subsection{Subgraph Reconstruction}

The reconstruction process of the evidence graph is depicted in Figure 3. We utilize networkx\footnote{The networkx can reorganize the list containing multiple groups of parent-child nodes into a graph. \url{https://networkx.org/}} to build two subgraphs using forward and reverse retrieval techniques. Reverse retrieval allows us to verify the accuracy of our findings. By intersecting the edges of the two subgraphs and removing any non-overlapping nodes and edges, we can construct a complete evidence graph. This evidence graph visually demonstrates the reasoning process from the initial question to the final answer.
\begin{equation}
G= \begin{cases}G_F \cup G_B & \text { if } \text{BSC}(G_F,G_B)>\gamma \\ G_B& \text { if } \text{BSC}(G_F,G_B)\leq  \gamma\end{cases}
\end{equation}

\noindent We first extract edges from the forward subgraph and the backward subgraph respectively, and then select by evaluating the BSC of the subgraph. If the BSC is less than $\gamma$, intersection of the bidirectional subgraphs is taken to reconstruct the graph, and new edges are not added twice for the existing nodes. 
\begin{equation}
B S C(G_F, G_B)=\frac{Edge_F \cap Edge_B}{Edge_F \cup Edge_B}
\end{equation}
where $Edge_F$,$Edge_B$ is the edge set of forward and backward subgraghs. If the BSC is greater than $\gamma$, the backward subgraph is reserved. A threshold value of $\gamma$ is used to determine whether the intersection of the two subgraphs should be used to construct the final evidence graph. The reason for using BSC and threshold value $\gamma$ is that, it can effectively improve the retrieval performance, by preserving the integrity and accuracy of the final evidence graph, also it can help to prevent from adding unnecessary edges.

\subsection{Answer Generation}
In order to generate an answer, the multiple evidences are fed into the reader as following. 

\begin{equation}
A = Reader_{T5} \left ( \left \{e_i | e_i \in G\right \}  \right ) 
\end{equation}

\noindent where evidences are ordered according to the structure of the retrieved evidence graph $G$. 
\par To measure the retrieval performance, We follow the previous settings \cite{JamesThorne2021DatabaseRO,zhu2022reasonchainqa} and use the classic T5 \cite{raffel2019exploring} model as the fixed reader, but this can easily be adapted to other pre-trained language models.

\section{Experimental}
In this section, we analyze the performance of different retrieval and reasoning systems on ReasonGraphQA, and investigate performance and limitations of our proposed graph-hop retrieval system.

\subsection{Compared Baselines}
We compare retrieval models of two retrieval mechanisms representative. Single-Hop retrieval method that retrieves all evidence at once (Random, BM25 \cite{Amati2009}, DPR \cite{karpukhin-etal-2020-dense}). Multi-hop retrieval methods retrieve one evidence iteratively in one step (GRR \cite{AkariAsai2019LearningTR}, MDR \cite{WenhanXiong2020AnsweringCO}, SSG \cite{JamesThorne2021DatabaseRO}). We use the code and parameter settings provided by the original papers for all baselines. For single-Hop retrieval models (BM25, DPR, SSG), we retrieve the top-k evidence, where k is the size of the golden evidence set.

We also explore the potential of large language models (LLM) in solving complex reasoning tasks through few-shot learning \cite{wei2022chain,weng2022large,weng2023neural}. To this end, we have developed five reasoning graph prompts for LLM, detailed in the appendix \ref{llm}. These prompts aim to enable the construction of a reason graph by LLM.
\noindent  

\begin{table*}[h]
\begin{center}
\renewcommand\arraystretch{1.3}
\resizebox{0.9\linewidth}{!}{%
	\centering
   \begin{tabular}{cl|ccc|cccc|c}
    \bottomrule \bottomrule
     \multicolumn{2}{c|}{\multirow{2}{*}{\textbf{Method}}}                                                                                  & \multicolumn{3}{c|}{\textbf{Explanation Graph}}     & \multicolumn{4}{c|}{\textbf{Evidence Set}}            & \multicolumn{1}{c}{\textbf{QA EM}}       \\                                             \multicolumn{2}{c|}{}&  GM$\uparrow$ &GS$\uparrow$ & GED$\downarrow$  &  F1$\uparrow$  &Precision$\uparrow$ &Recall $\uparrow$   &  EM$\uparrow$  &Acc$\uparrow$     \\ \hline
    \multirow{3}{*}{\shortstack{Single-Hop\\ Retrieval}}&\multicolumn{1}{l|}{Random} & -& -& - &-&-&13.292& - &37.509    \                                                                            \\ 
     &BM25 \cite{Amati2009}& -& -& - &-&-&70.842& - &62.423        \\
     &DPR \cite{karpukhin-etal-2020-dense}
& -& -& - &-&-&88.040& - &67.393            \\ \hline
       \multirow{3}{*}{\shortstack{Multi-Hop\\ Retrieval}}&GRR \cite{AkariAsai2019LearningTR}&24.915&25.187&5.862&71.447&\textbf{99.387}&60.132&25.051  &55.276              \\
     &SSG \cite{JamesThorne2021DatabaseRO} &34.717&35.12&6.437&75.806&78.233&77.935&53.846  &63.036                \\ 

     &MDR \cite{WenhanXiong2020AnsweringCO}&25.459&25.459&5.815&84.716&\underline{97.958}&79.971&62.832&51.259\\ \hline

\multirow{3}{*}{\shortstack{LLMs'\\ Retrieval}}&GPT-3 \cite{brown2020language}&0.070&17.135&8.043&12.752&23.973&10.624&0.070  &37.121              \\

     &GLM \citep{zeng2022glm}&0.680&4.762&7.095&11.163&21.259&8.676&0.680&38.023\\
          &Instruct-GPT \citep{LongOuyang2022TrainingLM} &35.908&54.767&\textbf{1.177}&71.898&67.466&81.788&40.779  & 56.489 \\ 
          \hline

\multirow{3}{*}{\shortstack{Graph-Hop\\ Retrieval}}& Graph-Hop's Forward &27.706&28.863&6.715&\underline{88.775}&\underline{87.202}&\underline{92.806}&\underline{67.120}&\underline{69.707} \\
&Graph-Hop's Backward & \underline{56.569}&\underline{57.862}&4.635&85.671&85.116&88.551 &64.057&67.597\\
     
&\multicolumn{1}{l|}{BGR (Bidirectional)}&\textbf{56.705}&\textbf{58.475}&\underline{4.703}&\textbf{91.809}&90.785&\textbf{95.227}&\textbf{68.822}&\textbf{70.184} \\ \bottomrule  \bottomrule
        \multicolumn{2}{c|}{\textbf{Human Bound}}   &92.152&93.154&0.181&98.134&98.731&97.544&96.412&95.125 \\\bottomrule  \bottomrule
    \end{tabular}}
    \caption{Main experimental results of BGR compared with three types of Retrieval methods on Retrieval-Reader architecture. In addition, we report the results of human in the test set to show the upper bound of human.}
	\label{3}
			\vspace{-0.3cm}
   \end{center}
\end{table*}
All methods are tested in the Dev set at the end of each round, and the model with the highest retrieval accuracy in the Dev set is selected for testing. We repeate the process three times by replacing the random seeds and average them as the final result.

\subsection{Implementation}
To measure retrieval mechanism in a fairer open-domain setting, We uniformly use T5 model \cite{raffel2019exploring} \footnote{\url{https://huggingface.co/t5-base}} as reader, and input retrieval evidence of different methods into a fine-tuned T5 model to generate answer. Specifically, we provide the correct evidence and questions in the training set to the reader for training, three readers were trained by different random seeds. A bert-base-uncased model is chosen as text encoder for extracting feature. We use AdamW \cite{IlyaLoshchilov2018DecoupledWD} with warm-up as the optimizer. The learning rate, epoch and batch size are set to $1\times10^{-5}$, 20, 8 respectively. Text maximum length $n$ was set as 30 and the $d$ was set as 768. 
\subsection{Evaluation Metrics}
In retrieval task, correctness were measured in terms of Explanation Graph and Evidence Set. Following previous works \cite{ZhilinYang2018HotpotQAAD,BhavanaDalvi2021ExplainingAW}, Exact Match (EM) , Precision, Recall and F1 was adopted. As for Explanation Graph evaluation, we used three indicators, Graph Matching (GM) evaluates whether the retrieved evidence graph is consistent with golden evidence graph. Graph Structure (GS) evaluates whether retrieved graph structure and golden graph structure are isomorphic, it will ignore nodes accuracy. Graph Editing Distance (GED) \cite{ZeinaAbuAisheh2015AnEG} measures how many steps does converting retrieved evidence graph to the golden one need. Then we use EM to measure the performance of QA task.

\subsection{Results and Analysis}

\begin{table}[h]

\begin{center}
\vspace{0.2cm}

\renewcommand\arraystretch{1.5}
\resizebox{6.5cm}{!}{%
\begin{tabular}{cccc}
\specialrule{.12em}{0pt} {0pt}
  Model & W/O Reason Graph & With Reason Graph \\\midrule
  GPT-3  & {\large 1.05} & {\large \textbf{23.55}} \\ 
  Instruct-GPT  & {\large 12.43} & {\large \textbf{45.15}}  \\
  GLM  & {\large 4.46}&{\large \textbf{7.15}}\\
  \specialrule{.12em}{0pt} {0pt} \\
\end{tabular}
}
\end{center}
      \vspace{-0.25cm}
  \caption{The Zero-shot performance of large language model (GPT-3: {\tt{}code-davinci-001} Instruct-GPT: {\tt{}code-davinci-002}) in ReasonGraphQA. We use a method similar to Chain of Thought to add the diagram structure to the input of LLM. See Appendix \ref{llm} for details.}
  \vspace{-0.5cm}
  \label{table3}
\end{table}

\textbf{The graph structure and the set retrieval both play a critical role.} As shown in Table \ref{3}, single-hop methods like DPR perform well in set recall and QA, while multi-hop methods like SSG excel in graph accuracy and QA. This highlights the importance of both the evidence graph structure and set retrieval for accurate question answering. This suggests that previous datasets \cite{qi-etal-2021-answering}, which only evaluate the accuracy of the retrieved set, are not sufficient for measuring QA performance. Additionally, as Table \ref{table3} shows, incorporating graph structure information into evidence results can significantly improve QA performance when using large language models. 



\textbf{LLM is capable of constructing inference diagrams.} In our LLM retrieval, as shown in Table \ref{3}, we discovered that while LLM has a low accuracy rate for the evidence set, it surpasses existing multi-hop retrieval in constructing inference graphs (especially for Instruct-GPT, Graph reasoning ability is close to Graph-Hop) which illustrates the reasoning potential of LLMs, which may be an important direction of future Graph-Hop research.

\textbf{Graph-Hop is more appropriate for ReasonGraphQA.} We note that multi-hop retrieval systems have high precision but low recall, as true nodes at the same level are ignored when retrieving along one reasoning chain. However, BGR can improve recall to 95.227\% by utilizing a bidirectional retrieval architecture. Additionally, Graph-Hop's Forward is better in evidence retrieval, while Backward has a higher graph construction capability. In the next section, we will further analyze Graph-Hop's performance and explain why BGR's performance is better after subgraph reconstruction.


\subsection{Ablation Study} 
\noindent\textbf{Bidirectional Retrieval}. To better understand cooperation mechanism of Forward retrieval and the Backward retrieval. We perform ablation study on retrieval direction. In Table \ref{3} we can clearly find that backward retrieval has a higher performance in the explanation graph, and forward retrieval has a higher performance in evidence retrieval. The BGR has better performance in explanation graph task, evidence retrieval task. And BGR outperform both forward and backward in QA task. This shows that bidirectional subgraph reconstruction(BCD algorithm) can make up deficiency of both and achieve a balance. 

\begin{figure}[t]
	\centering
	\includegraphics[scale=0.28]{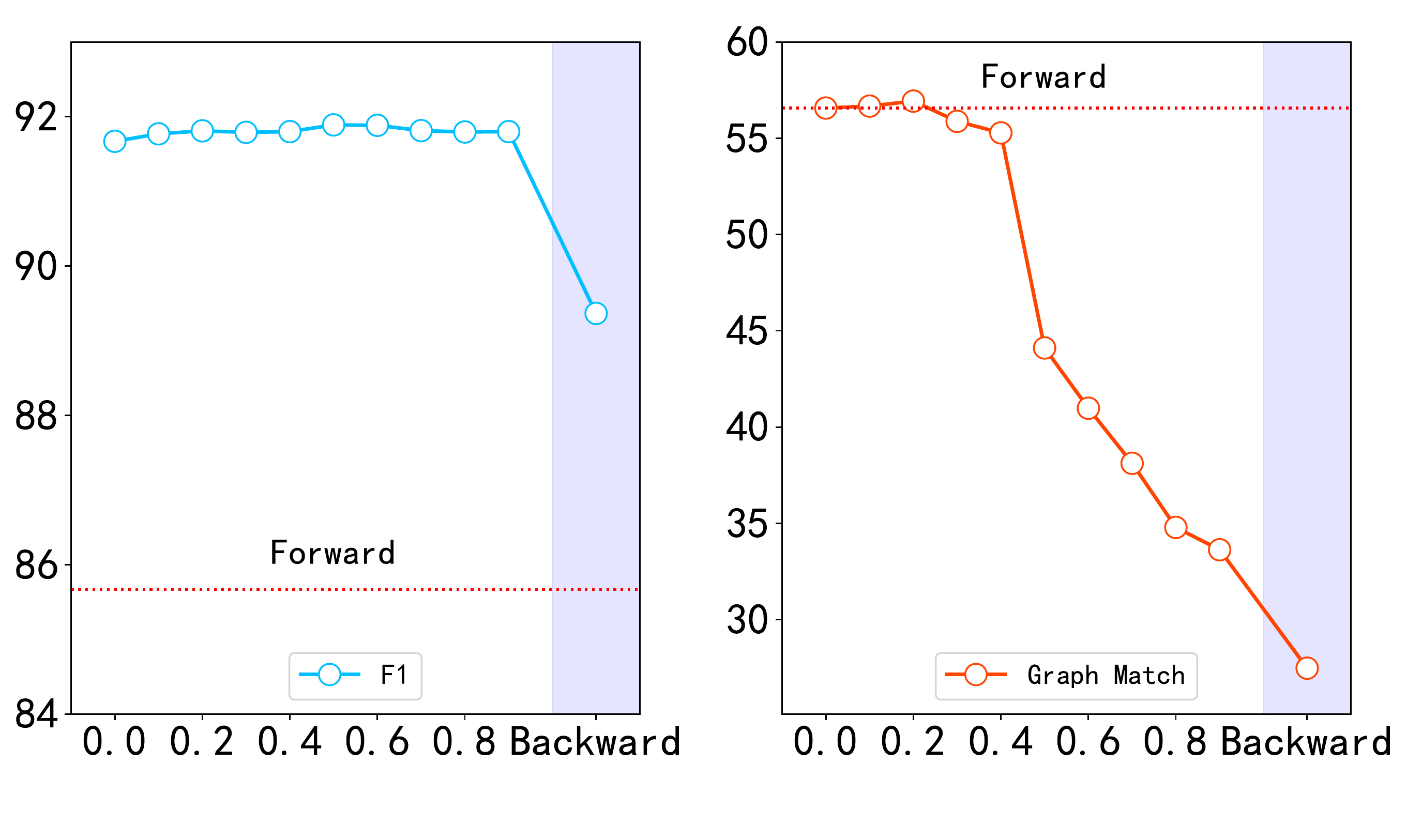}
	
	\vspace{-0.3cm}
	    \caption{F1 and GM changes with different $\gamma$.}
\vspace{-0.6cm}
	    \label{gamma}
\end{figure}
 \noindent\textbf{Bidirectional BGR with balanced $ \gamma $ value performs best}. As depicted in figure \ref{gamma}, we analyzed the effect of the value $ \gamma $ on the accuracy of retrieving the evidence set and graph. When $ \gamma $=1, the final evidence graph is $G_ B$. When $ \gamma=0 $, the evidence graph of samples that BSC $ \neq 0 $ is $G_ F \cup G_ B$. We found that the accuracy of bidirectional BGR is higher than that of forward and backward BGR, because $G_F$ performs better in graph structure, while $G_B$ tends to retrieve more accurate evidence sets, and the introduction of $ \gamma $ achieves a balanced result in the evidence set and graph structure. 

While BGR has achieve strong performance, its still an on-going challenge for graph-hop QA task. This is a meaningful task that are expected to promote development of TQA in knowledge reasoning and interpretability. 

\subsection{Futher Analysis on ReasonGraphQA.}
\begin{figure}[h]
	\centering
	\includegraphics[scale=0.175]{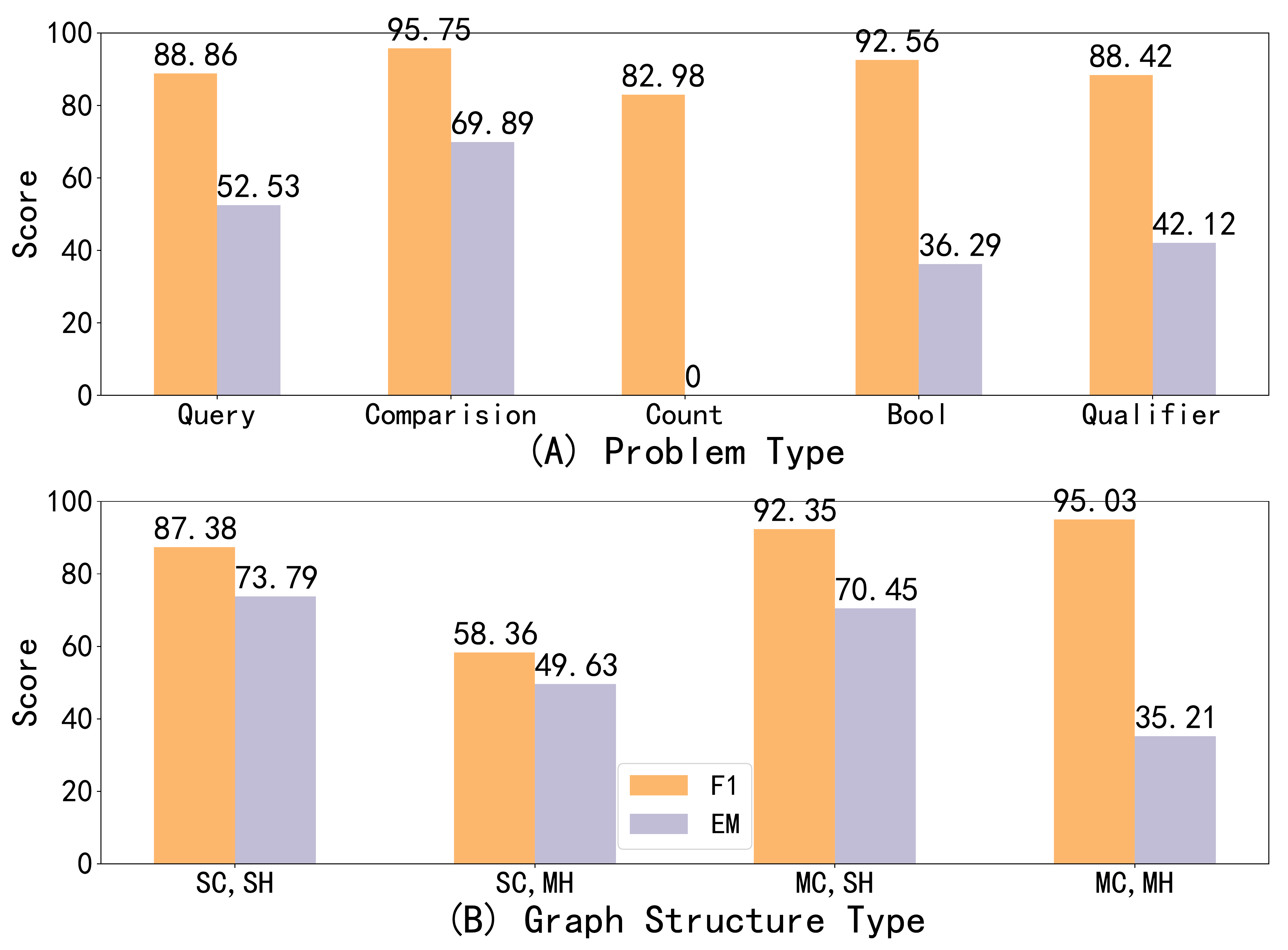}
	\vspace{-0.37cm}
	    \caption{The retrieval performance for different question and graph structure types.}

	    \label{cs}
\end{figure}

\noindent\textbf{BGR adapts to different question types.} We divide the test set into 5 different question types. Figure \ref{cs}(A) shows detailed accuracy of We can find that the evidence retrieval ability of the BGR can adapt to different kinds of questions, especially ``Comparison'' and ``Bool''. However, when faced with the task of constructing evidence graph, it is easy to miss nodes and edges. Even in the ``Count'' question, the BGR cannot correctly predict any explanation graph. This proves that the graph construction task still has a certain complexity, and the BGR still has a large room for improvement in the construction of retrieval evidence graphs.

\noindent\textbf{BGR performs well in complex, multi-hop explanation graph structures.} We classify and compare according to the graph structure, which are single-chain single-hop, single-chain multi-hop, multi-chain single-hop, and multi-chain multi-hop. In Figure \ref{cs}, more complex structure graph show the better retrieval performance, which proves that BGR can efficiently retrieve evidence in complex text question answering. In addition, BGR has achieved the best performance in MCMH explanation graph structures compared with the other three types, which even close to the QA accuracy with perfect retrieval. It shows that BGR is suitable for graph-hop retrieval. However, the more complex the graph structure is, the more edges there are. We believe that the modeling between edges is challenging due to the high similarity of edges between different nodes, which encourages researchers to conduct further research on explanation graph retrieval in the future. More detailed experimental results are provided in \ref{der}.

\begin{table}[h]
\begin{tabular}{ccc}
\begin{minipage}{0.17\textwidth}
\includegraphics[width=\textwidth]{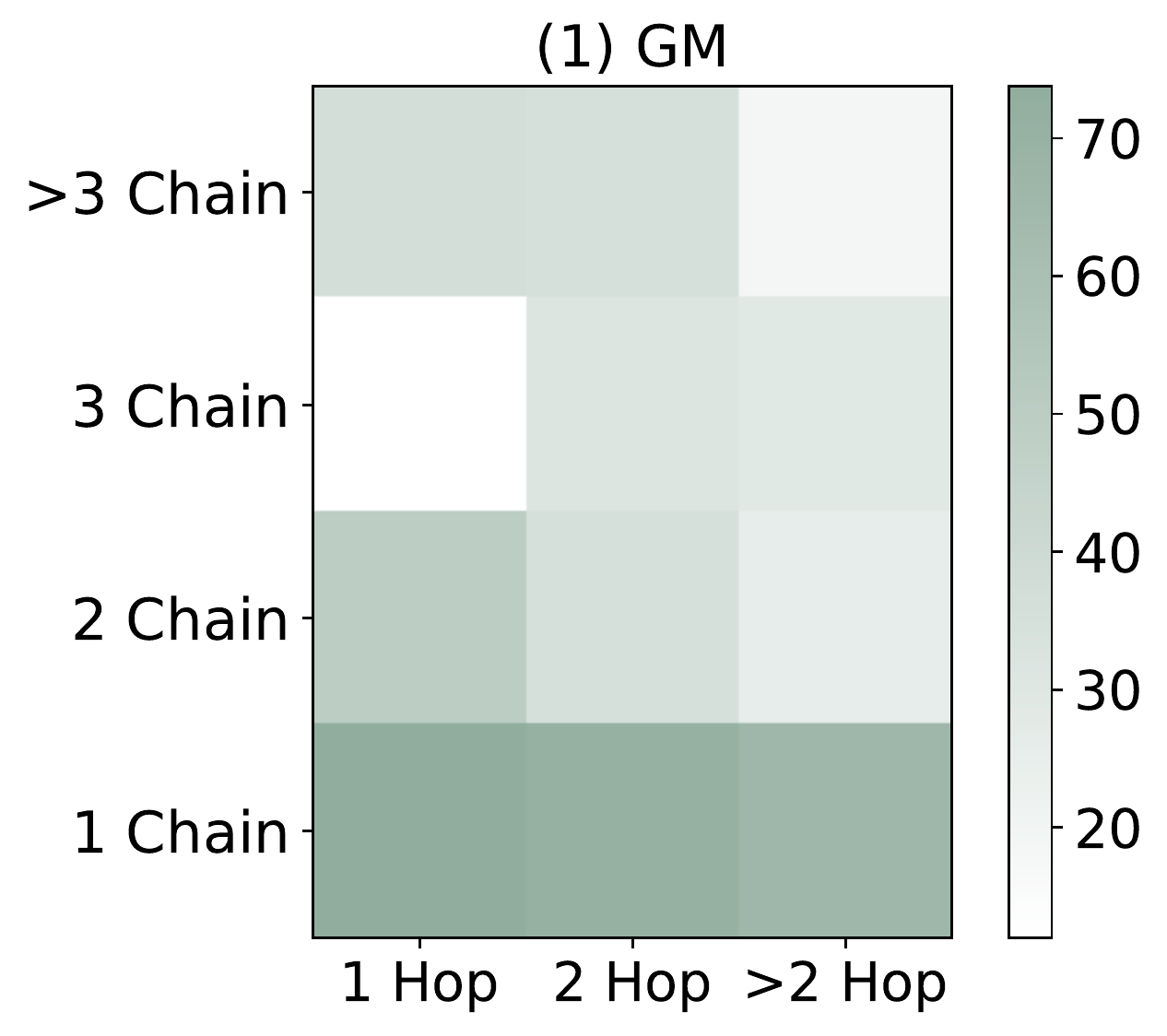}
\end{minipage} &
\begin{minipage}{0.125\textwidth}
\includegraphics[width=\textwidth]{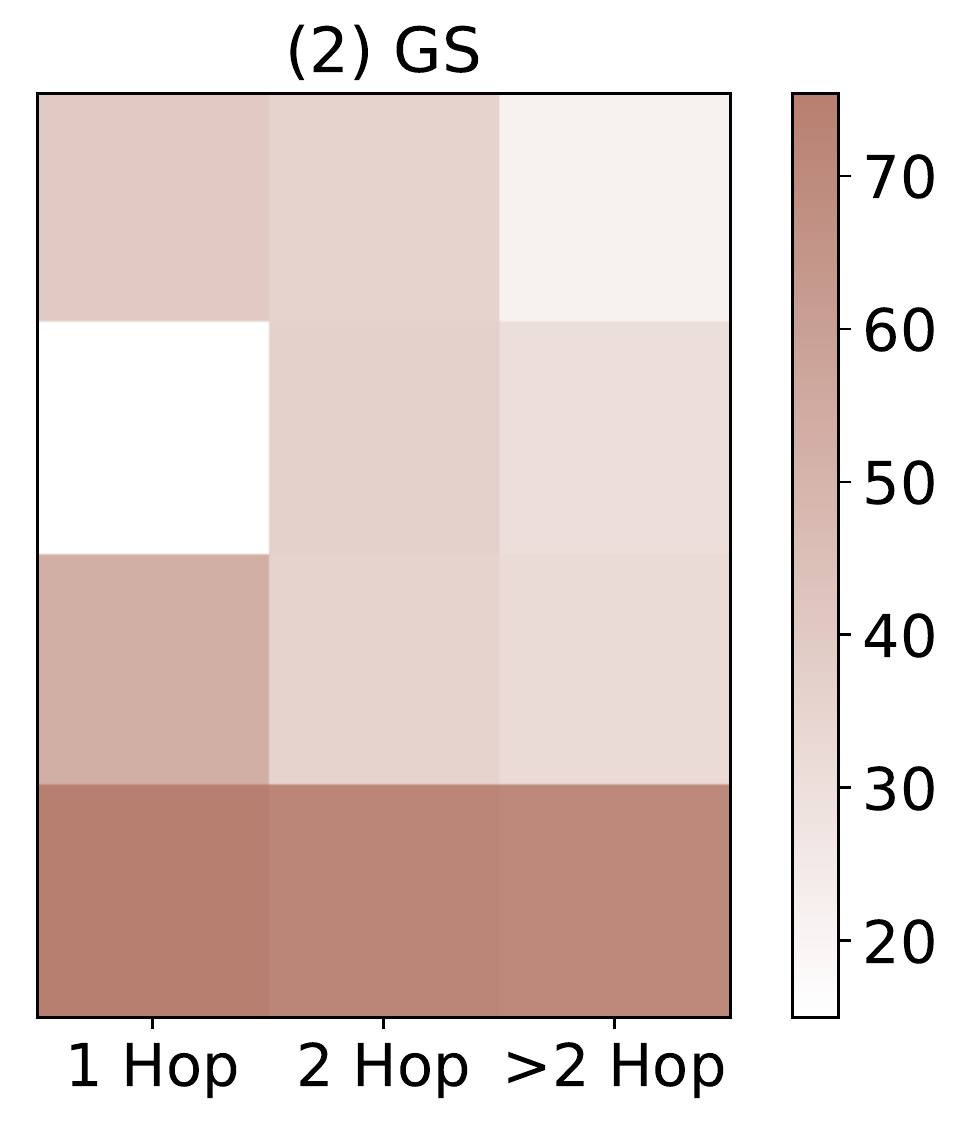}
\end{minipage} &
\begin{minipage}{0.125\textwidth}
\includegraphics[width=\textwidth]{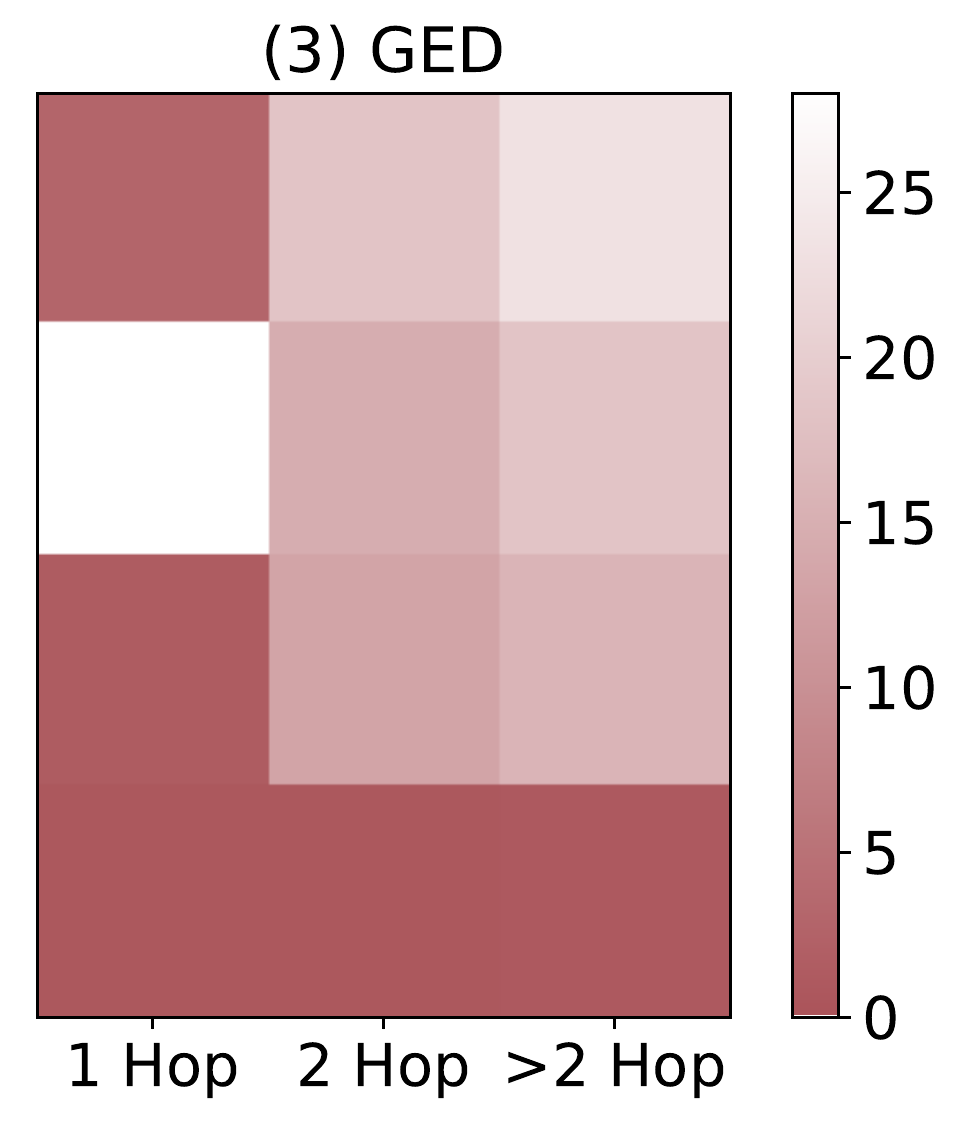}
\end{minipage} \end{tabular} 

\begin{tabular}{ccc} 
\begin{minipage}{0.17\textwidth}
\includegraphics[width=\textwidth]{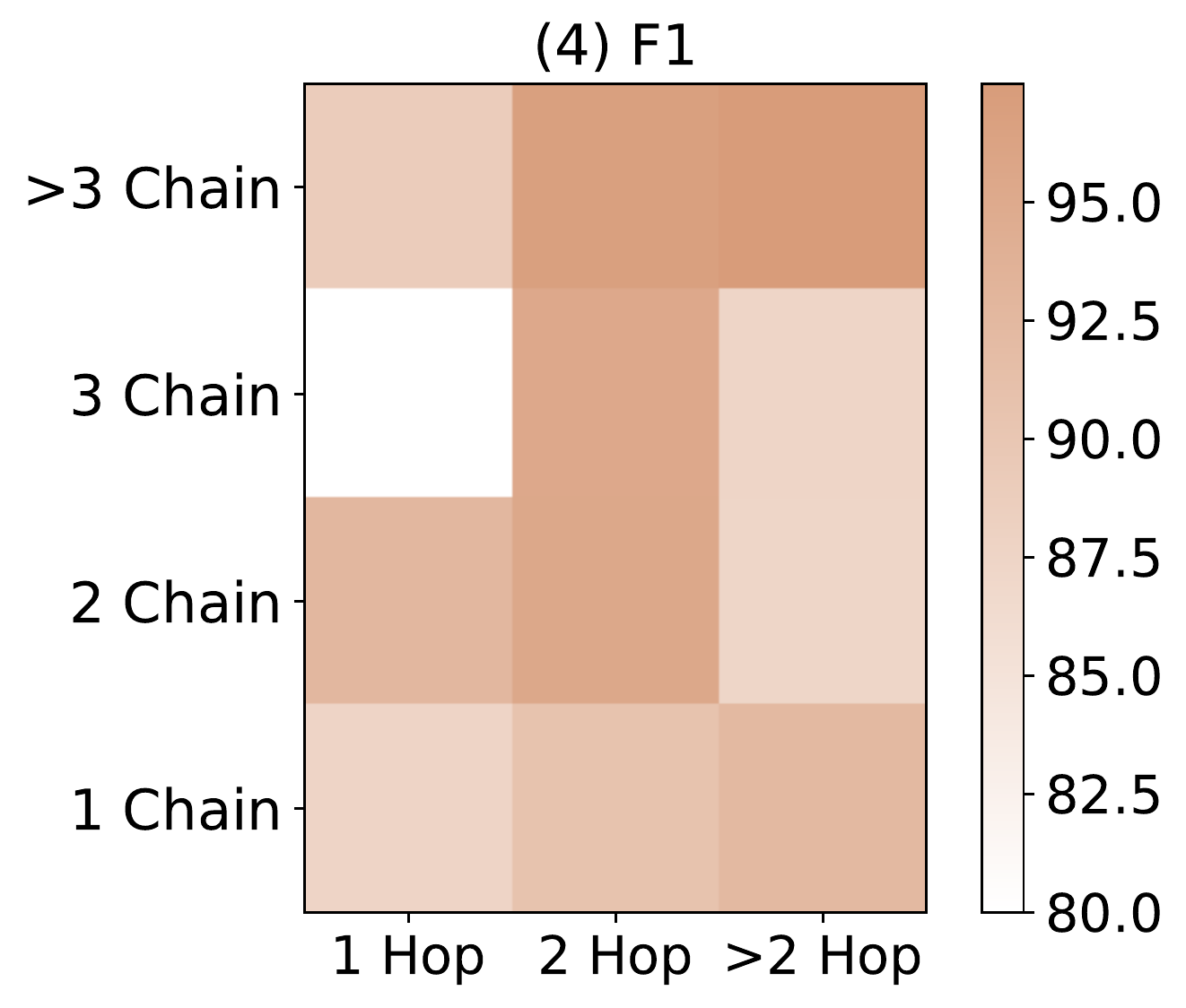}
\end{minipage} &
\begin{minipage}{0.125\textwidth}
\includegraphics[width=\textwidth]{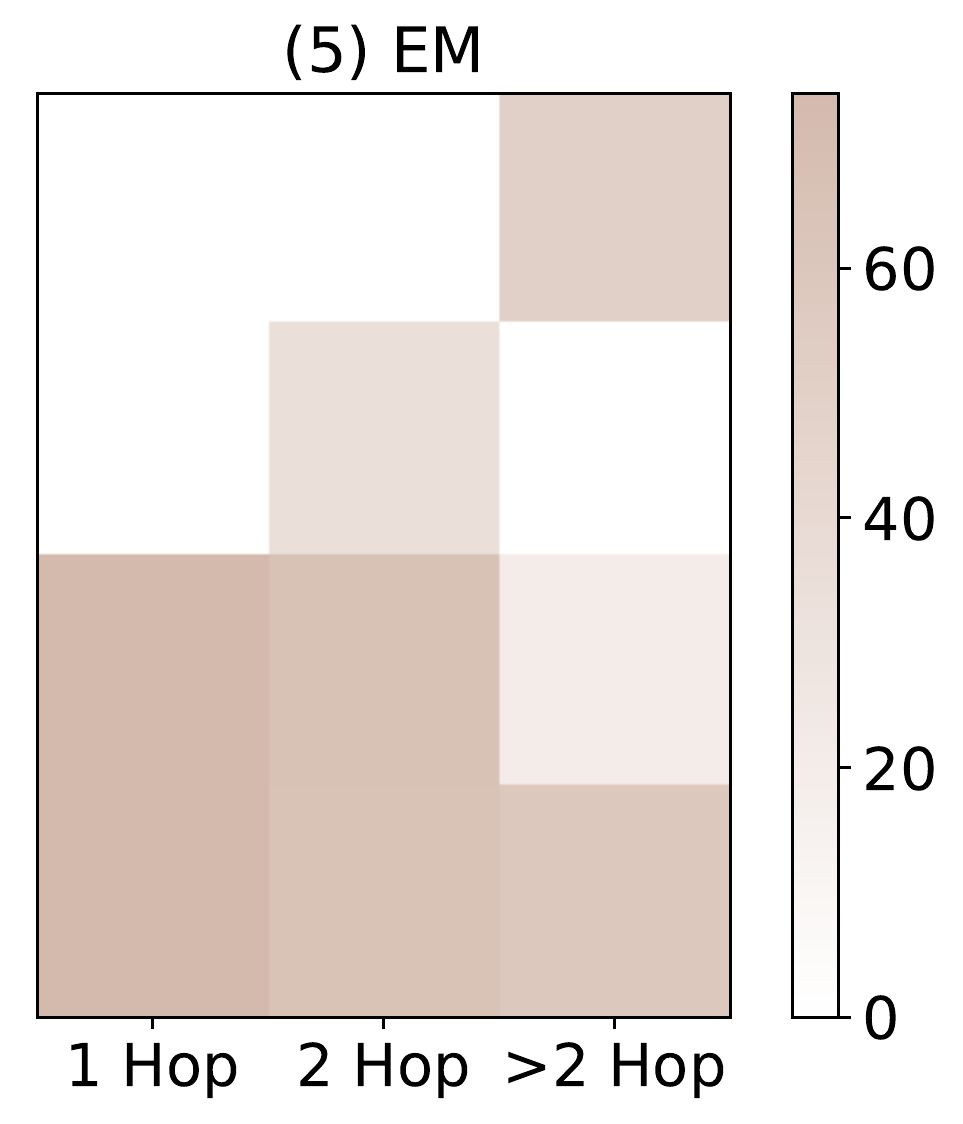}
\end{minipage} &
\begin{minipage}{0.125\textwidth}
\includegraphics[width=\textwidth]{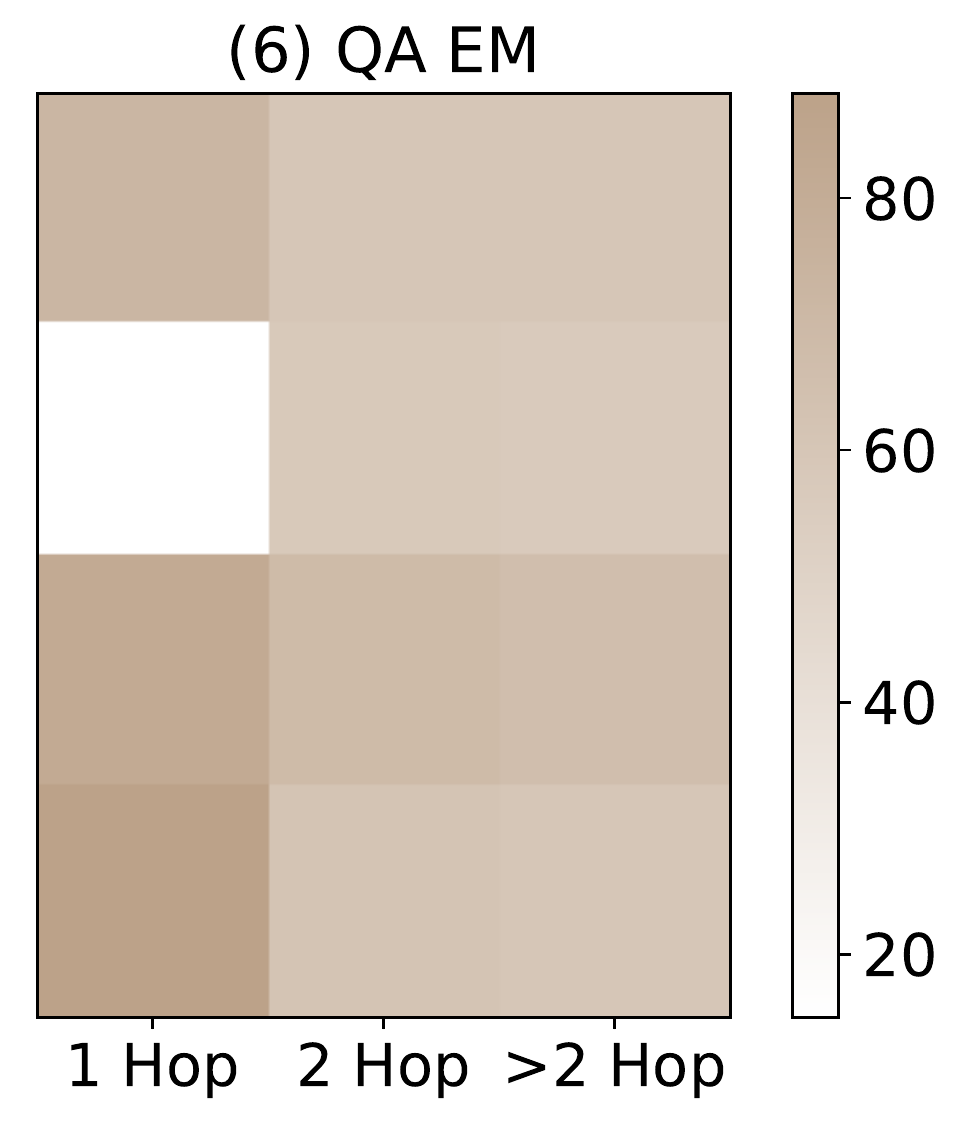}
\end{minipage}
\end{tabular}
\caption{Experimental results of BGR at different hops and different chain numbers.}
\label{hops and chains}
\end{table}

\noindent\textbf{The construction of multi-chain and multi-hop explanation graph is still challenging.} 
We have evaluated how varying hop and chain number of evidence graph structure influencing  graph structure (GM, GS, GED), evidence set (F1, EM), and question answering (QA EM). Our findings reveal that retrieving evidence graphs and answering questions from more complex evidence structures remains a challenging task. Specifically, as shown in Figure \ref{hops and chains}, the graph structure performance of evidence graph retrieval is strong for simple graphs but poor for complex ones, and the Exact Match of evidence sets retrieval is poor in complex graph structures. This results in relatively lower performance in question answering for complex graph structure samples.


\section{Conclusion}
Our study introduces the ReasonGraphQA dataset, the first textual database QA dataset with an explanation graph, which provides complex structured retrieval assistance for graph retrieval systems. We have tested various traditional evidence retrieval methods on the ReasonGraphQA dataset and evaluated them manually. Additionally, we propose the graph-hop retrieval paradigm and develop a bidirectional graph retrieval model, which significantly improves the evidence retrieval and graph construction capabilities of complex question answering by reconstructing reasoning paths in different directions. Future research utilizing the ReasonGraphQA dataset can enable fine-grained analysis of the explanation graph output from models, leading to further advancements in real and complex QA environments. While the current methods have several limitations, This presents opportunities for future research to improve upon them.

\section*{Limitations}

There are several limitations to our study. Firstly, the ReasonGraphQA dataset is built using a pre-trained language model to convert triples into a text database, which may lead to slight differences from the actual evidence. Secondly, we found that some complex graphs may have multiple possible explanation graphs, which can affect the model's training. We have provided detailed statistics on the quality of the ReasonGraphQA dataset in the supplementary material. Thirdly, the bidirectional graph retrieval model (BGR) has a higher time and space complexity compared to other methods, as it needs to retrieve both breadth and depth. This may affect its performance in pure multi-hop tasks.


\section*{Ethical Considerations}

Our research aimed to graph-hop retrieval in complex textual question answering. We use the pre-training language model to generate a large number of fluent evidences based knowledge base triples. we also realized that, due to our extensive use of pre-trained models with fact data from the Internet, the proposed method does not need manual annotation and reduce the carbon costs, it may be produce inappropriate text (For example, offensive, racially or gender-sensitive responses). 

We have carefully considered the above issues and provided the following details: (1) All fact data used is collected from the Internet, and it is inevitable that offensive, racially or gender-sensitive evidence facts will occur. We delete the sentences of evidence facts that are offensive, racially or gender-sensitive as much as possible. (2) The quality of the ReasonGraphQA dataset will affect the credibility of the robustness evaluation. We hope to maximize the reliability and implementability of the system based on such evaluation benchmarks. (3)  Finally, since the external knowledge base and KQA-Pro dataset are used to build the ReasonGraphQA, the information sources of these data also suffer from issues such as risk and bias. Reducing these potential risks requires ongoing research.

\bibliography{custom}

\begin{thebibliography}{37}
\expandafter\ifx\csname natexlab\endcsname\relax\def\natexlab#1{#1}\fi

\bibitem[{Abu-Aisheh et~al.(2015)Abu-Aisheh, Raveaux, Ramel, and
  Martineau}]{ZeinaAbuAisheh2015AnEG}
Zeina Abu-Aisheh, Romain Raveaux, Jean-Yves Ramel, and Patrick Martineau. 2015.
\newblock An exact graph edit distance algorithm for solving pattern
  recognition problems.
\newblock \emph{international conference on pattern recognition applications
  and methods}.

\bibitem[{Agarwal et~al.(2021)Agarwal, Ge, Shakeri, and
  Al-Rfou}]{agarwal-etal-2021-knowledge}
Oshin Agarwal, Heming Ge, Siamak Shakeri, and Rami Al-Rfou. 2021.
\newblock \href {https://doi.org/10.18653/v1/2021.naacl-main.278} {Knowledge
  graph based synthetic corpus generation for knowledge-enhanced language model
  pre-training}.
\newblock In \emph{Proceedings of the 2021 Conference of the North American
  Chapter of the Association for Computational Linguistics: Human Language
  Technologies}, pages 3554--3565, Online. Association for Computational
  Linguistics.

\bibitem[{Amati(2009)}]{Amati2009}
Giambattista Amati. 2009.
\newblock \href {https://doi.org/10.1007/978-0-387-39940-9_921} {\emph{BM25}},
  pages 257--260. Springer US, Boston, MA.

\bibitem[{Andor et~al.(2019)Andor, He, Lee, and
  Pitler}]{DanielAndor2019GivingBA}
Daniel Andor, Luheng He, Kenton Lee, and Emily Pitler. 2019.
\newblock Giving bert a calculator: Finding operations and arguments with
  reading comprehension.
\newblock \emph{empirical methods in natural language processing}.

\bibitem[{Asai et~al.(2020)Asai, Hashimoto, Hajishirzi, Socher, and
  Xiong}]{AkariAsai2019LearningTR}
Akari Asai, Kazuma Hashimoto, Hannaneh Hajishirzi, Richard Socher, and Caiming
  Xiong. 2020.
\newblock Learning to retrieve reasoning paths over wikipedia graph for
  question answering.
\newblock In \emph{International Conference on Learning Representations}.

\bibitem[{Brown et~al.(2020)Brown, Mann, Ryder, Subbiah, Kaplan, Dhariwal,
  Neelakantan, Shyam, Sastry, Askell et~al.}]{brown2020language}
Tom Brown, Benjamin Mann, Nick Ryder, Melanie Subbiah, Jared~D Kaplan, Prafulla
  Dhariwal, Arvind Neelakantan, Pranav Shyam, Girish Sastry, Amanda Askell,
  et~al. 2020.
\newblock Language models are few-shot learners.
\newblock \emph{Advances in neural information processing systems},
  33:1877--1901.

\bibitem[{Chen et~al.(2017)Chen, Fisch, Weston, and
  Bordes}]{DanqiChen2017ReadingWT}
Danqi Chen, Adam Fisch, Jason Weston, and Antoine Bordes. 2017.
\newblock \href {https://doi.org/10.18653/v1/P17-1171} {Reading {W}ikipedia to
  answer open-domain questions}.
\newblock In \emph{Proceedings of the 55th Annual Meeting of the Association
  for Computational Linguistics (Volume 1: Long Papers)}, pages 1870--1879,
  Vancouver, Canada. Association for Computational Linguistics.

\bibitem[{Dalvi et~al.(2021)Dalvi, Jansen, Tafjord, Xie, Smith, Pipatanangkura,
  and Clark}]{BhavanaDalvi2021ExplainingAW}
Bhavana Dalvi, Peter Jansen, Oyvind Tafjord, Zhengnan Xie, Hannah Smith,
  Leighanna Pipatanangkura, and Peter Clark. 2021.
\newblock Explaining answers with entailment trees.
\newblock \emph{empirical methods in natural language processing}.

\bibitem[{Dua et~al.(2019)Dua, Wang, Dasigi, Stanovsky, Singh, and
  Gardner}]{DheeruDua2019DROPAR}
Dheeru Dua, Yizhong Wang, Pradeep Dasigi, Gabriel Stanovsky, Sameer Singh, and
  Matt Gardner. 2019.
\newblock Drop: A reading comprehension benchmark requiring discrete reasoning
  over paragraphs.
\newblock \emph{north american chapter of the association for computational
  linguistics}.

\bibitem[{Gao et~al.(2021)Gao, Yao, and Chen}]{gao-etal-2021-simcse}
Tianyu Gao, Xingcheng Yao, and Danqi Chen. 2021.
\newblock \href {https://doi.org/10.18653/v1/2021.emnlp-main.552} {{S}im{CSE}:
  Simple contrastive learning of sentence embeddings}.
\newblock In \emph{Proceedings of the 2021 Conference on Empirical Methods in
  Natural Language Processing}, pages 6894--6910, Online and Punta Cana,
  Dominican Republic. Association for Computational Linguistics.

\bibitem[{Gupta et~al.(2019)Gupta, Lin, Roth, Singh, and
  Gardner}]{NitishGupta2019NeuralMN}
Nitish Gupta, Kevin Lin, Dan Roth, Sameer Singh, and Matt Gardner. 2019.
\newblock Neural module networks for reasoning over text.
\newblock \emph{Learning}.

\bibitem[{Jhamtani and Clark(2020)}]{jhamtani-clark-2020-learning}
Harsh Jhamtani and Peter Clark. 2020.
\newblock \href {http://arxiv.org/abs/2010.03274} {Learning to explain:
  Datasets and models for identifying valid reasoning chains in multihop
  question-answering}.

\bibitem[{Joshi et~al.(2017)Joshi, Choi, Weld, and
  Zettlemoyer}]{joshi-etal-2017-triviaqa}
Mandar Joshi, Eunsol Choi, Daniel Weld, and Luke Zettlemoyer. 2017.
\newblock \href {https://doi.org/10.18653/v1/P17-1147} {{T}rivia{QA}: A large
  scale distantly supervised challenge dataset for reading comprehension}.
\newblock In \emph{Proceedings of the 55th Annual Meeting of the Association
  for Computational Linguistics (Volume 1: Long Papers)}, pages 1601--1611,
  Vancouver, Canada. Association for Computational Linguistics.

\bibitem[{Karpukhin et~al.(2020)Karpukhin, Oğuz, Min, Lewis, Wu, Edunov, Chen,
  and tau Yih}]{karpukhin-etal-2020-dense}
Vladimir Karpukhin, Barlas Oğuz, Sewon Min, Patrick Lewis, Ledell Wu, Sergey
  Edunov, Danqi Chen, and Wen tau Yih. 2020.
\newblock \href {http://arxiv.org/abs/2004.04906} {Dense passage retrieval for
  open-domain question answering}.

\bibitem[{Loshchilov and Hutter(2018)}]{IlyaLoshchilov2018DecoupledWD}
Ilya Loshchilov and Frank Hutter. 2018.
\newblock Decoupled weight decay regularization.
\newblock In \emph{International Conference on Learning Representations}.

\bibitem[{Lu et~al.(2020)Lu, Abrego, Ma, Ni, and Yang}]{lu-etal-2021-multi}
Jing Lu, Gustavo~Hernandez Abrego, Ji~Ma, Jianmo Ni, and Yinfei Yang. 2020.
\newblock \href {http://arxiv.org/abs/2010.12523} {Neural passage retrieval
  with improved negative contrast}.

\bibitem[{Mou et~al.(2021)Mou, Yu, Chang, Feng, Zhang, and
  Su}]{XiangyangMou2021ComplementaryEI}
Xiangyang Mou, Mo~Yu, Shiyu Chang, Yufei Feng, Li~Zhang, and Hui Su. 2021.
\newblock Complementary evidence identification in open-domain question
  answering.
\newblock \emph{conference of the european chapter of the association for
  computational linguistics}.

\bibitem[{Ouyang et~al.(2022)Ouyang, Wu, Jiang, Almeida, Wainwright, Mishkin,
  Zhang, Agarwal, Slama, Ray, Schulman, Hilton, Kelton, Miller, Simens, Askell,
  Welinder, Christiano, Leike, and Lowe}]{LongOuyang2022TrainingLM}
Long Ouyang, Jeff Wu, Xu~Jiang, Diogo Almeida, Carroll Wainwright, Pamela
  Mishkin, Chong Zhang, Sandhini Agarwal, Katarina Slama, Alex Ray, John
  Schulman, Jacob Hilton, Fraser Kelton, Luke Miller, Maddie Simens, Amanda
  Askell, Peter Welinder, Paul Christiano, Jan Leike, and Ryan Lowe. 2022.
\newblock Training language models to follow instructions with human feedback.

\bibitem[{Paszke et~al.(2019)Paszke, Gross, Massa, Lerer, Bradbury, Chanan,
  Killeen, Lin, Gimelshein, Antiga, Desmaison, Kopf, Yang, DeVito, Raison,
  Tejani, Chilamkurthy, Steiner, Fang, Bai, and
  Chintala}]{NEURIPS2019_bdbca288}
Adam Paszke, Sam Gross, Francisco Massa, Adam Lerer, James Bradbury, Gregory
  Chanan, Trevor Killeen, Zeming Lin, Natalia Gimelshein, Luca Antiga, Alban
  Desmaison, Andreas Kopf, Edward Yang, Zachary DeVito, Martin Raison, Alykhan
  Tejani, Sasank Chilamkurthy, Benoit Steiner, Lu~Fang, Junjie Bai, and Soumith
  Chintala. 2019.
\newblock \href
  {https://proceedings.neurips.cc/paper/2019/file/bdbca288fee7f92f2bfa9f7012727740-Paper.pdf}
  {Pytorch: An imperative style, high-performance deep learning library}.
\newblock In \emph{Advances in Neural Information Processing Systems},
  volume~32. Curran Associates, Inc.

\bibitem[{Qi et~al.(2021)Qi, Lee, Sido, and Manning}]{qi-etal-2021-answering}
Peng Qi, Haejun Lee, Tg~Sido, and Christopher Manning. 2021.
\newblock \href {https://doi.org/10.18653/v1/2021.emnlp-main.292} {Answering
  open-domain questions of varying reasoning steps from text}.
\newblock In \emph{Proceedings of the 2021 Conference on Empirical Methods in
  Natural Language Processing}, pages 3599--3614, Online and Punta Cana,
  Dominican Republic. Association for Computational Linguistics.

\bibitem[{Raffel et~al.(2019)Raffel, Shazeer, Roberts, Lee, Narang, Matena,
  Zhou, Li, and Liu}]{raffel2019exploring}
Colin Raffel, Noam Shazeer, Adam Roberts, Katherine Lee, Sharan Narang, Michael
  Matena, Yanqi Zhou, Wei Li, and Peter~J Liu. 2019.
\newblock Exploring the limits of transfer learning with a unified text-to-text
  transformer.
\newblock \emph{arXiv preprint arXiv:1910.10683}.

\bibitem[{Rudra et~al.(2021)Rudra, Fernando, and Anand}]{KoustavRudra2021AnIA}
Koustav Rudra, Zeon~Trevor Fernando, and Avishek Anand. 2021.
\newblock An in-depth analysis of passage-level label transfer for contextual
  document ranking.
\newblock \emph{arXiv: Information Retrieval}.

\bibitem[{Shi et~al.(2022)Shi, Cao, Pan, Xiang, Hou, Li, Zhang, and He}]{KQA}
Jiaxin Shi, Shulin Cao, Liangming Pan, Yutong Xiang, Lei Hou, Juanzi Li,
  Hanwang Zhang, and Bin He. 2022.
\newblock Kqa pro: A dataset with explicit compositional programs for complex
  question answering over knowledge base.

\bibitem[{Thorne et~al.(2021)Thorne, Yazdani, Saeidi, Silvestri, Riedel, and
  Halevy}]{JamesThorne2021DatabaseRO}
James Thorne, Majid Yazdani, Marzieh Saeidi, Fabrizio Silvestri, Sebastian
  Riedel, and Alon Halevy. 2021.
\newblock Database reasoning over text.
\newblock In \emph{Proceedings of the 59th Annual Meeting of the Association
  for Computational Linguistics and the 11th International Joint Conference on
  Natural Language Processing (Volume 1: Long Papers)}, pages 3091--3104.

\bibitem[{Wei et~al.(2022)Wei, Wang, Schuurmans, Bosma, Chi, Le, and
  Zhou}]{wei2022chain}
Jason Wei, Xuezhi Wang, Dale Schuurmans, Maarten Bosma, Ed~Chi, Quoc Le, and
  Denny Zhou. 2022.
\newblock Chain of thought prompting elicits reasoning in large language
  models.
\newblock \emph{arXiv preprint arXiv:2201.11903}.

\bibitem[{Weng et~al.(2022)Weng, Zhu, He, Liu, and Zhao}]{weng2022large}
Yixuan Weng, Minjun Zhu, Shizhu He, Kang Liu, and Jun Zhao. 2022.
\newblock Large language models are reasoners with self-verification.
\newblock \emph{arXiv preprint arXiv:2212.09561}.

\bibitem[{Weng et~al.(2023)Weng, Zhu, Xia, Li, He, Liu, and
  Zhao}]{weng2023neural}
Yixuan Weng, Minjun Zhu, Fei Xia, Bin Li, Shizhu He, Kang Liu, and Jun Zhao.
  2023.
\newblock \href {http://arxiv.org/abs/2304.01665} {Neural comprehension:
  Language models with compiled neural networks}.

\bibitem[{Wolf et~al.(2020)Wolf, Debut, Sanh, Chaumond, Delangue, Moi, Cistac,
  Rault, Louf, Funtowicz, Davison, Shleifer, von Platen, Ma, Jernite, Plu, Xu,
  Scao, Gugger, Drame, Lhoest, and Rush}]{wolf-etal-2020-transformers}
Thomas Wolf, Lysandre Debut, Victor Sanh, Julien Chaumond, Clement Delangue,
  Anthony Moi, Pierric Cistac, Tim Rault, Rémi Louf, Morgan Funtowicz, Joe
  Davison, Sam Shleifer, Patrick von Platen, Clara Ma, Yacine Jernite, Julien
  Plu, Canwen Xu, Teven~Le Scao, Sylvain Gugger, Mariama Drame, Quentin Lhoest,
  and Alexander~M. Rush. 2020.
\newblock \href {https://www.aclweb.org/anthology/2020.emnlp-demos.6}
  {Transformers: State-of-the-art natural language processing}.
\newblock In \emph{Proceedings of the 2020 Conference on Empirical Methods in
  Natural Language Processing: System Demonstrations}, pages 38--45, Online.
  Association for Computational Linguistics.

\bibitem[{Wolfson et~al.(2020)Wolfson, Geva, Gupta, Gardner, Goldberg, Deutch,
  and Berant}]{TomerWolfson2020BreakID}
Tomer Wolfson, Mor Geva, Ankit Gupta, Matt Gardner, Yoav Goldberg, Daniel
  Deutch, and Jonathan Berant. 2020.
\newblock Break it down: A question understanding benchmark.
\newblock \emph{Transactions of the Association for Computational Linguistics}.

\bibitem[{Wu et~al.(2019)Wu, Zhang, and Feng}]{PeiyunWu2019ASO}
Peiyun Wu, Xiaowang Zhang, and Zhiyong Feng. 2019.
\newblock A survey of question answering over knowledge base.
\newblock \emph{China Conference on Knowledge Graph and Semantic Computing}.

\bibitem[{Xiong et~al.(2021)Xiong, Li, Iyer, Du, Lewis, Wang, Mehdad, Yih,
  Riedel, Kiela, and O{\u{g}}uz}]{WenhanXiong2020AnsweringCO}
Wenhan Xiong, Xiang~Lorraine Li, Srinivasan Iyer, Jingfei Du, Patrick Lewis,
  William~Yang Wang, Yashar Mehdad, Wen-tau Yih, Sebastian Riedel, Douwe Kiela,
  and Barlas O{\u{g}}uz. 2021.
\newblock Answering complex open-domain questions with multi-hop dense
  retrieval.
\newblock \emph{International Conference on Learning Representations}.

\bibitem[{Yang et~al.(2018)Yang, Qi, Zhang, Bengio, Cohen, Salakhutdinov, and
  Manning}]{ZhilinYang2018HotpotQAAD}
Zhilin Yang, Peng Qi, Saizheng Zhang, Yoshua Bengio, William~W. Cohen, Ruslan
  Salakhutdinov, and Christopher~D. Manning. 2018.
\newblock Hotpotqa: A dataset for diverse, explainable multi-hop question
  answering.
\newblock \emph{empirical methods in natural language processing}.

\bibitem[{Yuan et~al.(2021)Yuan, Neubig, and Liu}]{yuan2021bartscore}
Weizhe Yuan, Graham Neubig, and Pengfei Liu. 2021.
\newblock Bartscore: Evaluating generated text as text generation.
\newblock \emph{Advances in Neural Information Processing Systems}, 34.

\bibitem[{Zeng et~al.(2022)Zeng, Liu, Du, Wang, Lai, Ding, Yang, Xu, Zheng, Xia
  et~al.}]{zeng2022glm}
Aohan Zeng, Xiao Liu, Zhengxiao Du, Zihan Wang, Hanyu Lai, Ming Ding, Zhuoyi
  Yang, Yifan Xu, Wendi Zheng, Xiao Xia, et~al. 2022.
\newblock Glm-130b: An open bilingual pre-trained model.
\newblock \emph{arXiv preprint arXiv:2210.02414}.

\bibitem[{Zhang et~al.(2019)Zhang, Kishore, Wu, Weinberger, and
  Artzi}]{TianyiZhang2019BERTScoreET}
Tianyi Zhang, Varsha Kishore, Felix Wu, Kilian~Q. Weinberger, and Yoav Artzi.
  2019.
\newblock Bertscore: Evaluating text generation with bert.
\newblock \emph{Learning}.

\bibitem[{Zhu et~al.(2021)Zhu, Lei, Wang, Zheng, Poria, and
  Chua}]{FengbinZhu2021RetrievingAR}
Fengbin Zhu, Wenqiang Lei, Chao Wang, Jianming Zheng, Soujanya Poria, and
  Tat-Seng Chua. 2021.
\newblock Retrieving and reading: A comprehensive survey on open-domain
  question answering.
\newblock \emph{arXiv: Artificial Intelligence}.

\bibitem[{Zhu et~al.(2022)Zhu, Weng, He, Liu, and Zhao}]{zhu2022reasonchainqa}
Minjun Zhu, Yixuan Weng, Shizhu He, Kang Liu, and Jun Zhao. 2022.
\newblock Reasonchainqa: Text-based complex question answering with explainable
  evidence chains.
\newblock \emph{arXiv preprint arXiv:2210.08763}.

\end{thebibliography}
\bibliographystyle{acl_natbib}

\appendix

\section{Appendix}
\subsection{More Comprehensive Related Work \label{comprehensivework}}
Textual question answering (TQA) requires retrieve evidence from a large corpus to answer natural language questions. Some researchers proposed a novel TQA task over natural language database (NLDB) and support natural language database queries such as filtering, comparison and aggregation, where database is consist of unordered sets of textual facts \cite{JamesThorne2021DatabaseRO,zhu2022reasonchainqa}. Each fact is composed of text with different meanings rather than triples that unlike knowledge base QA. It requires comprehensive reasoning and retrieval of text sentences  \cite{TomerWolfson2020BreakID}. These NLDB tasks challenge the model with discretization and interpretable reasoning, where querying natural language databases with filtering, comparison, numerical operation queries \cite{NitishGupta2019NeuralMN,DheeruDua2019DROPAR}, and other operations \cite{DanielAndor2019GivingBA} still remains to be challenge.
 
Despite the rapid progress in TQA, they ignore the problem of multi-hop retrieval in multi-chain fact sets that may appear in complex textual question answering. For example, eQASC \cite{jhamtani-clark-2020-learning} and BeerQA \cite{qi-etal-2021-answering} are limited in breadth search, and the WIKINLDB \cite{JamesThorne2021DatabaseRO} are limited about depth search. In comparison, the proposed ReasonGraphQA requires graph retrieval from large-scale textual databases. And we focus on the discrete reasoning over textual evidences, which greatly evaluate the structured path modeling and discrete reasoning ability of QA systems over textual database.


\definecolor{c1}{HTML}{44cef6}
\definecolor{c2}{HTML}{c2d1e5}
\definecolor{c3}{HTML}{ab9350}
\definecolor{c4}{HTML}{02ad1c}
\definecolor{c5}{HTML}{9ae691}
\definecolor{c6}{HTML}{2b61db}
\definecolor{c7}{HTML}{8d4bbb}

\begin{table*}[h]
	\centering \small
   \begin{tabular}{m{1.5cm}<{\centering} m{2.5cm}<{\centering} m{5cm}<{\centering} m{1cm}<{\centering} m{4cm}<{\centering}}
   \hline
   \multicolumn{1}{c}{\multirow{1}{*}{ \textbf{Type}}}&
           \multicolumn{1}{c}{\textbf{Question}}   &  \textbf{Evidence}& \multicolumn{1}{c}{\multirow{1}{*}{ \textbf{Answer}}}  &   \textbf{Graph}   \\ \hline
           
                          \multicolumn{1}{c|}{\thead{\multirow{1}{*}{\begin{tabular}[c]{@{}c@{}}SigleChain \\SingleHop \end{tabular}}}}                   &   \begin{tabular}[l]{m{2.5cm}}How is \textcolor{orange}{Heaven's Gate} related to Joseph Cotten? \end{tabular}   &  \begin{tabular}[l]{m{5cm}}\textbf{$E$.}\textcolor{orange}{Heaven's Gate} stars Joseph Cotten. \\ ... \end{tabular}   &   \begin{tabular}[c]{m{1cm}}Cast member\end{tabular}&
       \begin{tikzpicture}
        \node[text centered](t){$Q$};
        \node[right = 1 of t, text centered](E00){$E$};
        \node[right = 2.4 of t](a){$A$};
        \draw[->, line width= 1] (t) -- node[above,font=\footnotesize]{}  (E00);
        \draw [->, line width= 1] (E00) -- node[above,font=\footnotesize]{Query}  (a);
        \end{tikzpicture}   \\ \hline

       \multicolumn{1}{c|}{\thead{\multirow{1}{*}{\begin{tabular}[c]{@{}c@{}}MultiChain \\SingleHop \end{tabular}}}}                     &  \begin{tabular}[l]{m{2.5cm}}Which one has more area between \textcolor{orange}{Billings} and \textcolor{c1}{Juneau}? \end{tabular}   &  \begin{tabular}[l]{m{5cm}}\textbf{$E^1$.}\textcolor{orange}{Billings} area is 113.467037. \\ \\ \textbf{$E^2$.}\textcolor{c1}{Juneau} area is 8427.626992. \\ ...\end{tabular}   & \begin{tabular}[c]{m{1cm}}  Juneau \end{tabular}&
       \begin{tikzpicture}
        \node[text centered](t){$Q$};
        \node[above right = 0.5 of t, text centered](E00){$E^1$};
        \node[below right= 0.5 of t, text centered](E01){$E^2$};
        \node[right = 1.8 of t](a){$A$};
        \draw[->, line width= 1] (t) -- node[above,font=\footnotesize]{}  (E00);
        \draw[->, line width= 1] (t) -- node[above,font=\footnotesize]{}  (E01);
        \draw [->, line width= 1] (E00) -- node[above right,font=\footnotesize]{Comparison}  (a);
        \draw [->, line width= 1] (E01) -- node[below right,font=\footnotesize]{Comparison}  (a);
        \end{tikzpicture}   \\ \hline

                       \multicolumn{1}{c|}{\thead{\multirow{1}{*}{\begin{tabular}[c]{@{}c@{}}SingleChain \\MultiHop \end{tabular}}}}                   &   \begin{tabular}[l]{m{2.5cm}}Is \textcolor{orange}{William}'s hometown \textcolor{c1}{the capital of the Netherlands}? \end{tabular}   &  \begin{tabular}[l]{m{5cm}}\textbf{$E_1$.}\textcolor{c1}{William} and his parents, \textcolor{c1}{their hometown} is \textcolor{c7}{Amsterdam}.\\ \\ \textbf{$E_2$.}\textcolor{c7}{Amsterdam} is located in the west of the Netherlands and is \textcolor{c1}{the capital of the Netherlands}. \\ ... \end{tabular}   & \begin{tabular}[c]{m{1cm}}  Continent\end{tabular}&       \begin{tikzpicture}
        \node[text centered](t){$Q$};
        \node[right = 0.5 of t, text centered](E0){$E_1$};
        \node[right = 0.5 of E0](E1){$E_2$};
        \node[right = 2.8 of t](a){$A$};
        \draw[->, line width= 1] (t) -- node[above,font=\footnotesize]{}  (E0);
        \draw[->, line width= 1] (E0) -- node[above,font=\footnotesize]{}  (E1);
        \draw[->, line width= 1] (E1) -- node[above,font=\footnotesize]{Bool}  (a);

        \end{tikzpicture}   \\ \hline

                                       \multicolumn{1}{c|}{\thead{\multirow{1}{*}{\begin{tabular}[c]{@{}c@{}}MultiChain \\MultiHop \end{tabular}}}}                   &  \begin{tabular}[l]{m{2.5cm}}Which city has larger \textcolor{orange}{population}, \textcolor{c7}{the capital of China} or \textcolor{c1}{the largest city in the United States}? \end{tabular}   &  \begin{tabular}[l]{m{5cm}}\textbf{$E_1^1$.} \textcolor{c7}{The capital of China} is \textcolor{c4}{Beijing}, which has a history of more than 3000 years  \\ \\ \textbf{$E_2^1$.}\textcolor{orange}{21.886 million people} live in \textcolor{c4}{Beijing}.
 \\ \\ \textbf{$E_1^2$.} \textcolor{c6}{New York} is \textcolor{c1}{the largest city in the United States}.  \\ \\ \textbf{$E_2^2$.}\textcolor{c6}{New York} has a \textcolor{orange}{large population of 8,510,000}  \\... \end{tabular}   & \begin{tabular}[l]{m{1cm}}  Quebec\end{tabular}&
       \begin{tikzpicture}
        \node[text centered](t){$Q$};
        \node[above right = 0.5 of t, text centered](E00){$E_1^1$};
        \node[right = 0.5 of E00](E01){$E_2^1$};
        \node[below right = 0.5 of t, text centered](E10){$E_1^2$};
        \node[right = 0.5 of E10](E11){$E_2^2$};
        \node[right = 2.8 of t](a){$A$};
        \draw[->, line width= 1] (t) -- node[above,font=\footnotesize]{}  (E00);
        \draw[->, line width= 1] (E00) -- node[above,font=\footnotesize]{}  (E01);
        \draw[->, line width= 1] (t) -- node[above,font=\footnotesize]{}  (E10);
        \draw[->, line width= 1] (E10) -- node[above,font=\footnotesize]{}  (E11);
        \draw[->, line width= 1] (E01) -- node[above right,font=\footnotesize]{Comparison}  (a);
        \draw[->, line width= 1] (E11) -- node[below right,font=\footnotesize]{Comparison}  (a);
        \end{tikzpicture}   \\ \hline

        ... & ... & ... & ... & ...  \\ \hline
        
    \end{tabular}
    \caption{Explanation graph example. We use the same color to represent the same nodes.}
	\label{cos}
			\vspace{-0.2cm}
\end{table*}

\begin{table}[t]
		\centering \small
		\renewcommand\arraystretch{1.2} \setlength{\tabcolsep}{1.4mm}
		\begin{tabular}{c|c}
			\noalign{\hrule height 1pt}

			Example&Data \\
			
						\hline
			 \multicolumn{1}{c|}{\thead{\multirow{2}{*}{\begin{tabular}[c]{@{}c@{}}Triplet \\ Facts \end{tabular}}}}&\multicolumn{1}{c}{\thead{\multirow{2}{*}{\begin{tabular}[c]{@{}c@{}}\{Amsterdam, location, Western of Netherlands\}, \\\{Amsterdam, the capital of, the Netherlandsp\} \end{tabular}}}}  \\
			 & \\
			 \hline
			 			 \multicolumn{1}{c|}{\thead{\multirow{3}{*}{\begin{tabular}[c]{@{}c@{}}Input \end{tabular}}}}&\multicolumn{1}{c}{\thead{\multirow{3}{*}{\begin{tabular}[c]{@{}c@{}}\textit{Please describe the following entities in} \\ \textit{one sentence:} \underline{Amsterdam, location, Western }\\ \underline{of Netherlands. the capital of, the Netherlandsp}  \end{tabular}}}}  \\
			 			 & \\
			 			 & \\
			 			 \hline
			 					 			 \multicolumn{1}{c|}{\thead{\multirow{3}{*}{\begin{tabular}[c]{@{}c@{}}Output \end{tabular}}}}&\multicolumn{1}{c}{\thead{\multirow{3}{*}{\begin{tabular}[c]{@{}c@{}}\textbf{Amsterdam is located } \\\textbf{ in the west of theNetherlands and is  } \\\textbf{the capital of the Netherlands.} \end{tabular}}}}  \\
			 			 & \\
			 			 & \\
			 			 \hline

			\noalign{\hrule height 1pt}

		\end{tabular}
		\caption{Example when using the pre-trained language model to generate facts}
		\vspace{-0.4cm}
		\label{Reasongraph}
	\end{table}

\label{Quality} 

\subsection{Data Analysis}
\label{sec:appendix_dataanalysis}
\begin{figure}[t]
\centering
\subfigure[Graph Structure Number] 
{\includegraphics[width=6cm]{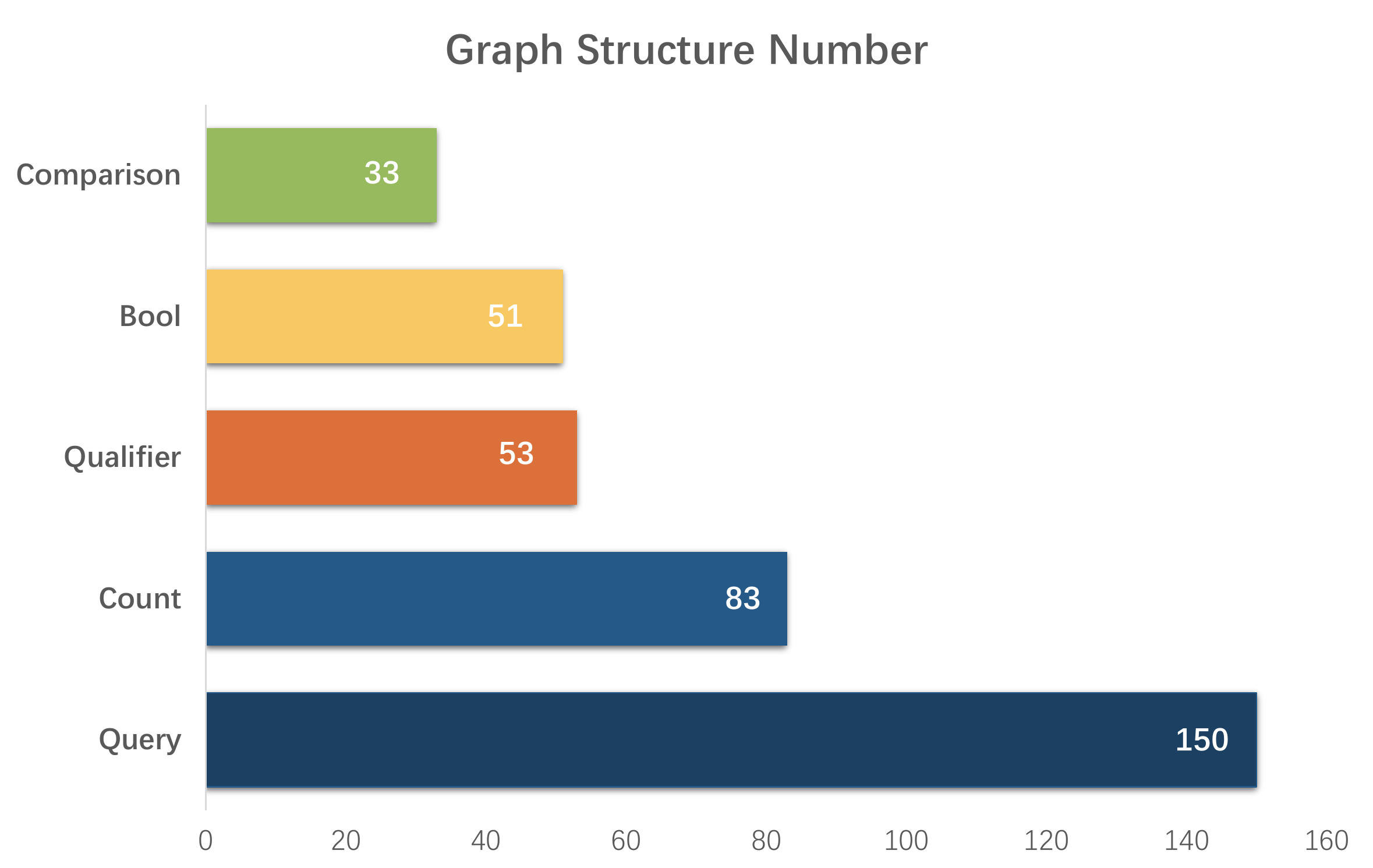}} 
\subfigure[Graph Distribution]{\includegraphics[width=6cm]{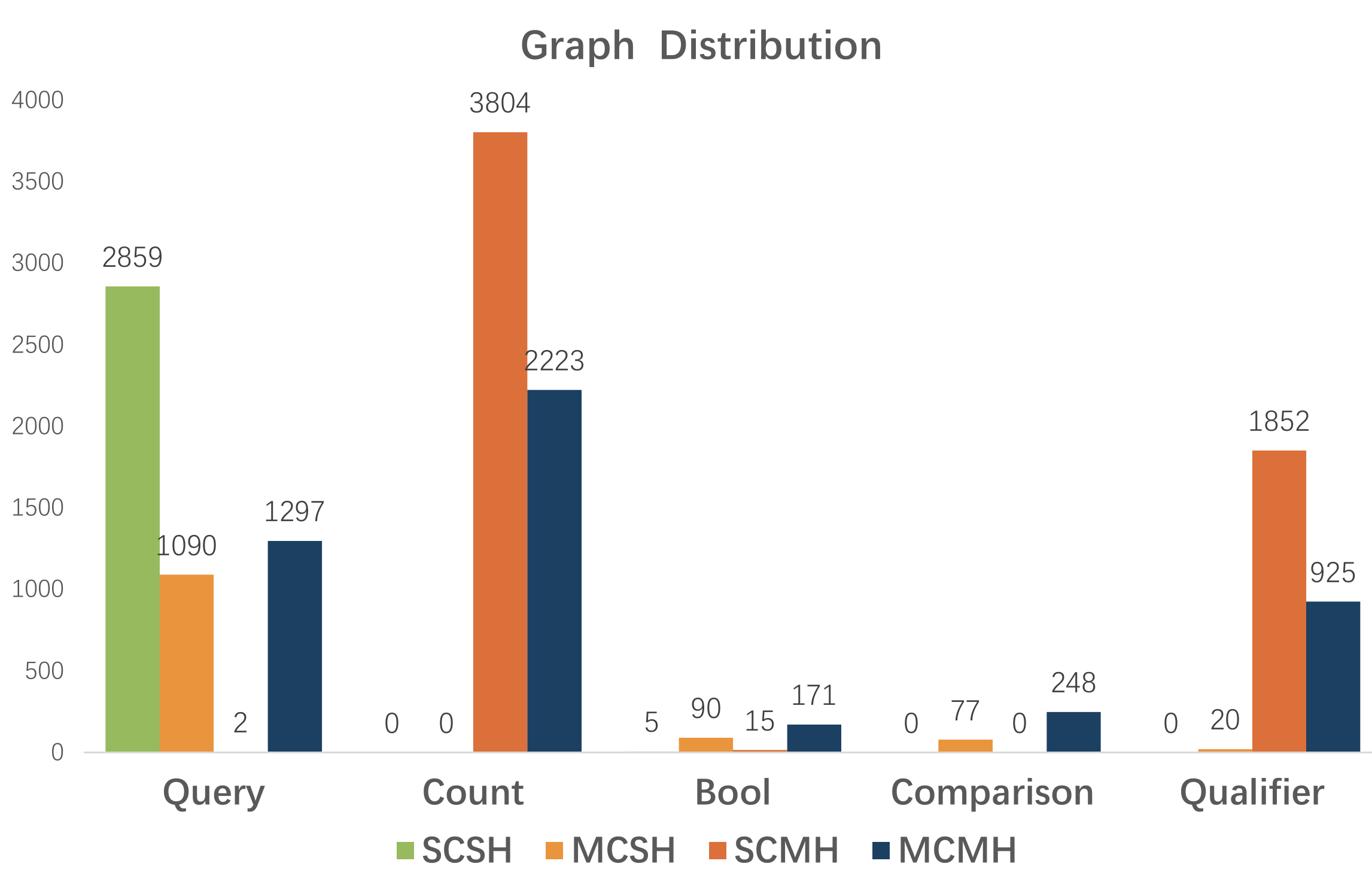}}
\caption{Graph statistics of ReasonGraphQA} 
\label{Graph statistics}  
\end{figure}

Each graph has an average of 5.3 edges and 4.2 nodes on everage. ReasonGraphQA contains 262 nonisomorphic graph structures, According to nine different ask strategies of KQA PRO, we divide questions into five types. in Figure \ref{Graph statistics}, which includes ``Query'', ``Comparison'', ``Count'', ``Bool'', ``Qualifier''. the ``Comparison'' involves comparison of multiple evidences. The ``Query'' type inquires head or tail entity of relational knowledge, the ``Qualifier'' query for attributes and relations, the ``Count'' type's answer is number, and the ``bool'' is to judge correctness of a statement. Most question type involves a variety of graph reasoning. The question types that involve the most graph structures are ``Queryname'', ``Count'' and ``Queryattribute'', which comprehensively involve value comparison, relational knowledge, and time knowledge. This further shows the complexity of our data set.

\subsection{Data Quality}

Figure \ref{f21} shows our quality assessment template. We engaged three graduate students to conduct manual evaluation of a randomly selected portion of our dataset and the test set. These students, hailing from China, were provided with a set of samples marked by the author as a reference for their evaluation. They conducted a thorough evaluation according to established standards, working an average of 10 days. Work eight hours a day. We compensated them for their efforts with a total of \$1750, which exceeds the average labor standard in China.

In Table \ref{5}, we can find that a large proportion of the samples only have some problems in the fluency of the evidence text and the problem text. 

We manually inspect randomly selected samples and assess its quality of questions, facts and evidence graphs by human. We randomly sampled 500 samples, covering the 10 graph structure types with the largest number of samples. For questions, the annotator is required to score according to fluency and comprehensibility. Of the 500 questions, only 9 are considered unsmooth. This is because there is a typo in the question. All the questions are understandable.

In order to evaluate the quality of mapping knowledge map triples to facts, we evaluated 200 text databases (each text database contains 25 facts) and scored them from smoothness, faithfulness and sufficiency (whether irrelevant information is included). When and only when all facts meet the requirements, the fact base is considered to meet the requirements. All evidences of 182/200 text databases are fluent, and only one or two of the remaining 18 text databases do not meet the requirements of fluency.

To evaluate the quality of the mapping of facts from knowledge triples, 500 sampled facts were scored based on smoothness, faithfulness, and sufficiency. 98.2\% (491/500) facts were smooth, with only 9 containing repeated text. 98.6\% (493/500) facts were faithful to the relation of the triples, with only 7 containing additional information. Of these, 3 fact replaced incorrect information with correct information, resulting in a faithfulness and sufficiency score of 0. The remaining 4 facts contained additional information that enriched the context. 

In addition to the separate evaluation of questions, facts, and evidence graphs, we also evaluated the overall quality of 100 randomly sampled databases (each containing 25 facts) using the six evaluation indicators mentioned above. Of these, 92 databases were deemed to be of good quality.

We believe that although some samples of ReasonGraphQA have problems, the overall score is high, and the average perfect sample score can reach 9.21, which reflects the high quality of our dataset.


\subsection{Performance of Graph-Hop when different graph structures}
\label{der}
Table \ref{160} presents a detailed analysis of the performance of Graph-Hop on four different types of graph structures. The results indicate that Graph-Hop demonstrates strong performance across all the structures tested. Among them, Graph-Hop particularly excels in its ability to navigate and understand multi-chain and multi-hop structures. These structures are known to be challenging for traditional graph traversal methods, making Graph-Hop's performance on these structures all the more noteworthy. Additionally, it should be noted that the results in Table 5 demonstrate that Graph-Hop is an effective and efficient tool for handling and understanding complex graph structures.

\begin{table}[h]
	\centering \small
   \begin{tabular}{c|ccc}
   \hline
    \textbf{Method} &\textbf{Training} & \textbf{Prediction} &  \textbf{EC}     \\ \hline
    Forward Retrieval & $4\pm1 $&$0.15\pm0.1$ &  $2.5\pm0.5$\\
    Backward Retrieval & $4\pm1$ &$0.1\pm0.1$ &  $2.5\pm0.5$\\
    BGR & $7.5\pm1 $&$0.25\pm0.1 $& $5\pm0.5$ \\
    \hline

    \end{tabular}
    \caption{Time (Hour) and Energy Consumption (KWH) statistics of Bidirectional Graph-hop Retrieval }
	\label{8}

\end{table}

	\begin{table*}[t]

\renewcommand\arraystretch{1.3}
	\centering \small
   \begin{tabular}{cl|ccc|cccc|c}
\bottomrule \bottomrule
    \multicolumn{1}{c|}{\multirow{2}{*}{\textbf{Graph}}} & \multicolumn{1}{c|}{\multirow{2}{*}{\textbf{Method}}}                                                                                  & \multicolumn{3}{c|}{\textbf{Explanation Graph}}     & \multicolumn{4}{c|}{\textbf{Retrieval}}            & \multicolumn{1}{c}{\textbf{QA EM}}       \\ 
    \multicolumn{1}{c|}{}&\multicolumn{1}{c|}{}                                                                                     &  GM$\uparrow$ &GA$\uparrow$ & GED$\downarrow$  &  F1$\uparrow$  &Precision$\uparrow$ &Recall $\uparrow$   &  EM$\uparrow$  &Acc$\uparrow$      \\  \midrule

                \multicolumn{1}{c|}{\thead{\multirow{3}{*}{\begin{tabular}[c]{@{}c@{}}SC\\SH\end{tabular}}}} 
            & Forward &\textbf{77.419}&\textbf{78.629}&0.633&\textbf{87.401}&\textbf{84.761}&93.548&\textbf{77.419}&\textbf{88.710}\\
                    \multicolumn{1}{c|}{}& Backward &72.581&73.790&\textbf{0.631}&85.645&82.305&92.742&72.581&87.903\\
                                                                 
        \multicolumn{1}{c|}{}& Bidirectional  &73.790&75.403&0.665&87.375&83.868&\textbf{95.161}&73.790&88.306\\\bottomrule

                         \multicolumn{1}{c|}{\thead{\multirow{3}{*}{\begin{tabular}[c]{@{}c@{}}MC\\SH\end{tabular}}}} &
             Forward &\textbf{50.370}&\textbf{53.333}&1.511&80.025&79.383&54.074&57.037&77.037\\
                    \multicolumn{1}{c|}{}& Backward &47.407&52.593&1.889&69.012&67.037&72.963&49.630&75.556\\
                                                                 
        \multicolumn{1}{c|}{}& Bidirectional  &49.630&52.593&\textbf{1.400}&\textbf{85.358}&\textbf{83.951}&\textbf{90.012}&\textbf{60.741}&\textbf{81.481}\\\bottomrule

        \multicolumn{1}{c|}{\thead{\multirow{3}{*}{\begin{tabular}[c]{@{}c@{}}SC\\MH\end{tabular}}}} &
             Forward &21.678&22.202&2.294&91.578&90.415&94.843&\textbf{73.776}&\textbf{62.413}\\
                    \multicolumn{1}{c|}{}&Backward &\textbf{70.629}&\textbf{71.678}&\textbf{0.675}&92.009&91.305&95.105&73.077&61.538\\
                                                                 
        \multicolumn{1}{c|}{}& Bidirectional  &70.455&71.678&0.680&\textbf{92.347}&\textbf{91.643}&\textbf{95.542}&72.727&61.364\\\bottomrule
         \multicolumn{1}{c|}{\thead{\multirow{3}{*}{\begin{tabular}[c]{@{}c@{}}MC\\MH\end{tabular}}}} & Forward &9.533&10.117&15.889&90.305&90.464&91.742&61.479&67.315\\
                    \multicolumn{1}{c|}{}&  Backward &\textbf{35.603}&\textbf{36.187}&\textbf{11.652}&83.004&84.332&83.328&\textbf{69.758}&62.451\\
                                                                 
        \multicolumn{1}{c|}{}& Bidirectional  &34.825&35.603&13.125&\textbf{95.055}&\textbf{94.913}&\textbf{96.351}&64.397&\textbf{68.289}\\\bottomrule \bottomrule
        
    \end{tabular}
    \caption{We show the experimental results under different graph structures.}
	\label{160}

\end{table*}

\subsection{Hyperparameter and Detailed Experimental Results}
\begin{table}[h]
	\centering \small
   \begin{tabular}{c|c}
   \hline
    \textbf{Hyperparameter} & \textbf{Value}         \\ \hline
    BGR$_{Encoder}$  & bert-base-uncased \\
   
    BGR$_{Decoder}$ & Transformers\\
    Hidden Size & 768 \\
    Num Layers & 12(Encoder) + 1(Decoder) \\
    $\gamma$ & 0.2 \\
    Dropout & 0.1\\
    Linear decay & 1e-6 \\
    Learning Rate & 1e-5\\
    Reader Learning Rate & 1e-4 \\
    Batch size & 8\\
    Max length & 30\\
    Num Epochs & 20\\
    GPU DRAM usage & 18G \\
    Params & 139M \\
    \hline
    \end{tabular}
    \caption{ Hyper-parameter settings.}
	\label{hps}
\end{table}
All our experiments were conducted in a 10900k CPU computer with 128G memory and RTX3090 GPU. We conduct experiments using the PyTorch \cite{NEURIPS2019_bdbca288} and the huggingface \cite{wolf-etal-2020-transformers} framework. We use linear decay of learning rate by $1\times10^{-6}$ and the Table \ref{hps} shows all our super parameter settings. We have counted the training time, prediction time and energy consumption in Table \ref{8} for the BGR model.

In Table \ref{Graph statistics}, we further evaluate the performance of the bidirectional  method in different graph structures. We divided the ReasonGraphQA dataset into four parts according to the graph structure. They are single-chain single-hop, multi-chain single-hop, single-chain multi-hop, and multi-chain multi-hop. This helps to understand and analyze the performance of the current model for different structures.

We can find that multi-chain and multi-hop tasks are more difficult than other tasks, whether in graph construction or retrieval. The bidirectional  method can help highlight the path representation advantage of bidirectional retrieval, and it can be significantly improved in multi-chain and multi-hop tasks. But it will slightly reduce the performance of the model for simple problems. We think that this is because for simple problems, the representations in bidirectional are mostly consistent, and there is no major conflict in the learning direction. Therefore, bidirectional  is difficult to significantly improve the effect.

\subsection{Large language Models Setting}
\label{llm}
We evaluated the performance of the original GPT-3 \cite{brown2020language} ({\tt{}code-davinci-001}) model, the Instruct-GPT model~\citep{LongOuyang2022TrainingLM} ({\tt{}code-davinci-002}), and GLM \cite{zeng2022glm} model on the ReasonGraphQA datasets. All GPT models' predictions were obtained through OpenAI's API. Due to server limitations, the GLM model used Int8 inference on 8 RTX3090 with 512G RAM Memory. 

\begin{figure}[t]
	\centering
	\includegraphics[scale=0.46]{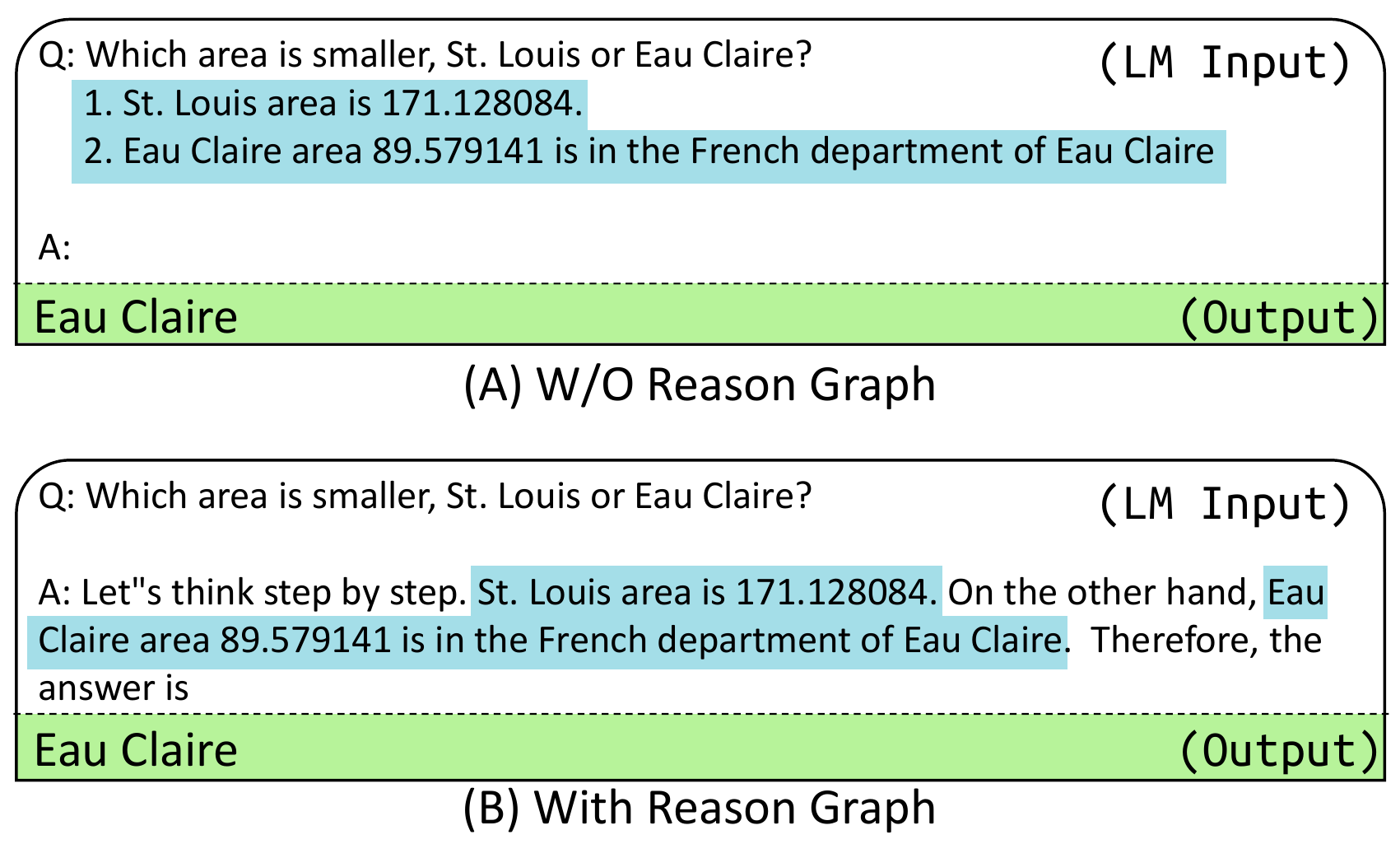}
    \caption{Two different comparison methods when using LLM for zero-shut QA.}
	\label{dataset}

\end{figure}

We conducted all experiments in the few-shot setting, without any fine-tuning the orginal language model. Apart from the context, we have not provided any other prompt text.

When incorporating graph structure into the input of a language model, the thought chain can serve as a useful approach. As depicted in Figure 6, we utilize the phrase ``Then'' to denote the relationship between adjacent nodes, and ``On the other hand'' to indicate the relationship between different chains.

\begingroup
\begin{table}[h]
    \centering
    \small

    \begin{tabular}{p{\linewidth}}
        \toprule
        \underline{\textbf{\textsc{Prompt for ReasonGraphQA}}} \\
        \vspace{-2mm}
        \textbf{Q:} Does the film Veronica Mars (whose release date is 2014-03-08) or Rambo (that succeeds Rambo III) have the shorter running time?\\ 
0. This Side of Resurrection is a drama film which was directed by Joaquim Sapinho and produced by the latter. The film was released on September 8, 2011.\\ 
1. The film Back on Track has a duration of +114 minutes.\\ 
2. Red Dust is a film that has a duration of +107 minutes.\\ 
3. The film Slack Bay was made in 2016 and has a 14 class in the Indie rating. It was directed by Bruno Dumont and stars Valeria Bruni Tedeschi. It was released on 26 January 2017.\\ 
4. The film Saving Mr. Banks was a feature film which was released in 2013 and 2014. Its title is for Hispanic America. The film stars Tom Hanks and Melanie Paxson.\\ 
5. Fierro is a Spanish language film which was released in 2007. It is a drama with Hector Calori and Roly Serrano as the main characters.\\ 
6. Me You Them is a film starring Lima Duarte and Regina Case. It was released on 16 May 2000 and 22 March 2001. The film's musical score was provided by Gilberto Gil.\\ 
7. The film Come What May has a duration of +114 minutes.\\ 
8. The biographical film La macchinazione was written and directed by David Grieco. It was released in 2016 and is a drama film with a 2017 release date.\\ 
9. The Rewrite is a 2014 film starring Allison Janney. The film has an AL rating. It was released on 13 November 2014 and on 25 December 2014.\\ 
10. The Rambo is a 90-minute television series. \\ 
11. The film Joyeux No l is set in 1914. Its stars are Rolando Villazón, Dany Boon and Joachim Bißmeier. The film was nominated for the Academy Award for Best International Feature Film.\\ 
12. The Italian film Open Your Eyes (1997) is about telepresence. Its main characters are Jorge de Juan and Isabel Serrano. The director of photography is Hans Burman. The film also stars Ion Gabella.\\ 
13. Robin Hood ( 1991 British film ) has a duration of +133 minute.\\ 
14. The film redoubtable has a duration of +107 minutes.\\ 
15. The film Summer Games, a drama, was released in 2011. Its director was Rolando Colla and stars Giorgio Gobbi and Alessia Barela.\\ 
16. Roland Verhavert is a director of films and is related to the category of films directed by Roland Verhavert.\\ 
17. No Retreat, No Surrender 2 is a 1987 film directed by Corey Yuen. It stars Max Thayer and Matthias Hues. The film was released on 28 January 1988.\\ 
18. The sequel to Rambo III was released in 2005. \\ 
19. Forbidden Hours is a drama film that was released in 1927 and 1928. It was directed by Harry Beaumont and starred Ramon Novarro.\\ 
20. Veronica Mars is 107 minutes long. \\ 
21. Humidity is a drama film from Serbia which was released on 15 February 2016. The film's cast includes Slaven Do lo.\\ 
22. Rango ( 2011 film ) has a duration of +107 minute.\\ 
23. The Wonders is a film which was released in 2014 and 2015. It stars Sabine Timoteo and Sam Louwyck.\\ 
24. Risc vs. Reward is an extended play by Photek. It was released in 1997 and its genre is downtempo. It was followed by Modus Operandi.\\ 
 \\
        \vspace{-1mm}
        \textbf{A:} Q-18-20;Q-18-10; \\
        \bottomrule
    \end{tabular}
        \caption{
    Few-shot exemplars 1.
    }
    \label{tab:appendix-aqua-prompt}
\end{table}
\endgroup
\begingroup
\begin{table}[h]
    \centering
    \small

    \begin{tabular}{p{\linewidth}}
        \toprule
        \underline{\textbf{\textsc{Prompt for ReasonGraphQA}}} \\
        \vspace{-2mm}
        \textbf{Q:} Does the film Veronica Mars (whose release date is 2014-03-08) or Rambo (that succeeds Rambo III) have the shorter running time?\\ 
0. This Side of Resurrection is a drama film which was directed by Joaquim Sapinho and produced by the latter. The film was released on September 8, 2011.\\ 
1. The film Back on Track has a duration of +114 minutes.\\ 
2. Red Dust is a film that has a duration of +107 minutes.\\ 
3. The film Slack Bay was made in 2016 and has a 14 class in the Indie rating. It was directed by Bruno Dumont and stars Valeria Bruni Tedeschi. It was released on 26 January 2017.\\ 
4. The film Saving Mr. Banks was a feature film which was released in 2013 and 2014. Its title is for Hispanic America. The film stars Tom Hanks and Melanie Paxson.\\ 
5. Fierro is a Spanish language film which was released in 2007. It is a drama with Hector Calori and Roly Serrano as the main characters.\\ 
6. Me You Them is a film starring Lima Duarte and Regina Case. It was released on 16 May 2000 and 22 March 2001. The film's musical score was provided by Gilberto Gil.\\ 
7. The film Come What May has a duration of +114 minutes.\\ 
8. The biographical film La macchinazione was written and directed by David Grieco. It was released in 2016 and is a drama film with a 2017 release date.\\ 
9. The Rewrite is a 2014 film starring Allison Janney. The film has an AL rating. It was released on 13 November 2014 and on 25 December 2014.\\ 
10. The Rambo is a 90-minute television series. \\ 
11. The film Joyeux No l is set in 1914. Its stars are Rolando Villazón, Dany Boon and Joachim Bißmeier. The film was nominated for the Academy Award for Best International Feature Film.\\ 
12. The Italian film Open Your Eyes (1997) is about telepresence. Its main characters are Jorge de Juan and Isabel Serrano. The director of photography is Hans Burman. The film also stars Ion Gabella.\\ 
13. Robin Hood ( 1991 British film ) has a duration of +133 minute.\\ 
14. The film redoubtable has a duration of +107 minutes.\\ 
15. The film Summer Games, a drama, was released in 2011. Its director was Rolando Colla and stars Giorgio Gobbi and Alessia Barela.\\ 
16. Roland Verhavert is a director of films and is related to the category of films directed by Roland Verhavert.\\ 
17. No Retreat, No Surrender 2 is a 1987 film directed by Corey Yuen. It stars Max Thayer and Matthias Hues. The film was released on 28 January 1988.\\ 
18. The sequel to Rambo III was released in 2005. \\ 
19. Forbidden Hours is a drama film that was released in 1927 and 1928. It was directed by Harry Beaumont and starred Ramon Novarro.\\ 
20. Veronica Mars is 107 minutes long. \\ 
21. Humidity is a drama film from Serbia which was released on 15 February 2016. The film's cast includes Slaven Do lo.\\ 
22. Rango ( 2011 film ) has a duration of +107 minute.\\ 
23. The Wonders is a film which was released in 2014 and 2015. It stars Sabine Timoteo and Sam Louwyck.\\ 
24. Risc vs. Reward is an extended play by Photek. It was released in 1997 and its genre is downtempo. It was followed by Modus Operandi.\\ 
 \\
        \vspace{-1mm}
        \textbf{A:} Q-6;Q-22-13; \\
        \bottomrule
    \end{tabular}
        \caption{
    Few-shot exemplars 2.
    }
    \label{tab:appendix-aqua-prompt}
\end{table}
\endgroup
\begingroup
\begin{table}[h]
    \centering
    \small

    \begin{tabular}{p{\linewidth}}
        \toprule
        \underline{\textbf{\textsc{Prompt for ReasonGraphQA}}} \\
        \vspace{-2mm}
        \textbf{Q:} Which area is smaller, St. Louis or Eau Claire?\\ 
0. The area of Belle Prairie City, Illinois is +0.45 square mile.\\ 
1. La Prairie , Illinois has an area of +0.23 square mile.\\ 
2. The area of Bellerive, Missouri is +0.873665 square kilometres.\\ 
3. Pierpont, in Missouri, has an area of +0.25 square mile.\\ 
4. Montclare, Chicago has an area of +2.56 square kilometres.\\ 
5. Fredericton is located in the area of 130680000. \\ 
6. Lewistown, Illinois has a surface area of +5.177494 square kilometres.\\ 
7. Old Shawneetown, Illinois has an area of +0.53 square mile.\\ 
8. St. Marys, Iowa has an area of +0.361422 square kilometres.\\ 
9. San Jose, Illinois has an area of +0.50 square mile.\\ 
10. La Loge Pas-de-Calais has an area of +0.68 square kilometres.\\ 
11. Eau Claire, Calgary has a surface area of +0.4 square kilometres.\\ 
12. Saunemin, Illinois has an area of +0.24 sq. mi.\\ 
13. St. Louis area is 171.128084. \\ 
14. Southside (East Chicago) covers an area of +1.0 sq km.\\ 
15. Eau Claire area 89.579141 is in the French department of Eau Claire. \\ 
16. The area of Southern View, Illinois is +0.52 square mile.\\ 
17. Des Arc, Missouri has a area of +0.551754 square kilometres.\\ 
18. The area of Van Buren, Missouri is +5.183903 square kilometres.\\ 
19. With an area of +12.135017 square kilometres, Dolton, Illinois, is in Illinois.\\ 
20. Huey, Illinois is a city with a total area of +0.22 sq. mi.\\ 
21. Lucerne, Missouri has an area of +0.648701 square kilometres.\\ 
22. The area of Lac de la Lauch is +0.11 square kilometre.\\ 
23. Iola, Illinois has a total area of +0.97 square miles.\\ 
24. Lambert, Missouri has an area of +0.133048 square kilometres.\\ 

 \\
        \vspace{-1mm}
        \textbf{A:} Q-13;Q-15; \\
        \bottomrule
    \end{tabular}
        \caption{
    Few-shot exemplars 3.
    }
    \label{tab:appendix-aqua-prompt}
\end{table}
\endgroup
\begingroup
\begin{table}[h]
    \centering
    \small

    \begin{tabular}{p{\linewidth}}
        \toprule
        \underline{\textbf{\textsc{Prompt for ReasonGraphQA}}} \\
        \vspace{-2mm}
        \textbf{Q:} Which work of Fritz Leiber Junior was awarded  Nebula Award for Best Novella?\\ 
0. Brian Aldiss won the Nebula Award for Best Novella.\\ 
1. Frederik Pohl won the Nebula Award for Best Novel in 1976 for Gateway and in 1977 for Man Plus.\\ 
2. James Tiptree Jr. won the Nebula Award for Best Novella for the novel "A Momentary Taste of Being". He also won the Nebula Award for Best Novella for "Houston, Houston, Do You Read?" in 1975. He also won the Nebula Award for Best Novella in 1976 and in 1985.\\ 
3. Fritz Leiber Junior won the Nebula Award for Best Novella for Ill Met in Lankhmar. \\ 
4. The Locus Award for Best Novel is a literary award. It was formerly known as the Locus Award for Best Science Fiction Novel.\\ 
5. Fritz Leiber is a human being who wrote science fiction. He was influenced by Robert E. Howard. He won the Hugo Award for Best Novelette. He was nominated for the Hugo Award for Best Dramatic Presentation. He won the Geffen Award.\\ 
6. David Gerrold was nominated for the Nebula Award for Best Novella in 1998.\\ 
7. The winner of the Nebula Award for Best Novelette is Eugie Foster.\\ 
8. Robert Silverberg was the winner of the Hugo Award for Best Novelette.\\ 
9. In addition to the Nebula Award nominations, Le Guin was nominated for the Nebula Award for Best Novel for Powers. \\ 
11. Then there is the Nebula Award for Best Novel for Powers, the book that was nominated for the Nebula Award for Best Novel.\\ 
13. Jonathan Lethem was nominated for the Nebula Award for Best Novella in 2000.\\ 
14. Geoff Ryman's novelette What We Found won the Nebula Award for Best Novelette in 2007. He was nominated for the award in 2011.\\ 
15. Fritz Leiber was nominated for the Hugo Award for Best Novel.\\ 
16. Fritz Leiber Junior won the Nebula Award for Best Novella. \\ 
17. Samuel R. Delany won the Nebula Award for Best Novel for The Einstein Intersection in 1966. He also won the Nebula Award for Best Novel for Dhalgren in 1967. He also won the Nebula Award for Best Novel for Triton in 1975.\\ 
18. Fritz Leiber won the Hugo Award for Best Novelette.\\ 
19. David Gerrold won the Nebula Award for Best Novelette.\\ 
20. The novel Binti won the Nebula Award for Best Novella in 2015.\\ 
21. Lucius Shepard won the Hugo Award for Best Novella for Barnacle Bill the Spacer in 1993. He has also won the Hugo Award for Best Novella in 1989, 1990, 2001, 2008 and 1987.\\ 
22. The Martian Child won the Hugo Award for Best Novelette in 1995. David Gerrold is the winner of the Hugo Award.\\ 
23. David Gerrold was nominated for the Nebula Award for Best Novel.\\ 
24. John Kessel won the Nebula Award for Best Novelette for Pride and Prometheus in 2008.\\ 

 \\
        \vspace{-1mm}
        \textbf{A:} Q-16;Q-3; \\
        \bottomrule
    \end{tabular}
        \caption{
    Few-shot exemplars 4.
    }
    \label{tab:appendix-aqua-prompt}
\end{table}
\endgroup
\begingroup
\begin{table}[h]
    \centering
    \small

    \begin{tabular}{p{\linewidth}}
        \toprule
        \underline{\textbf{\textsc{Prompt for ReasonGraphQA}}} \\
        \vspace{-2mm}
        \textbf{Q:} What is the connection between Margot Kidder to Brian De Palma?\\ 
0. The drama genre Counterblast was created by Guy Morgan and Jack Whittingham. It stars Martin Miller and Nova Pilbeam. Margaretta Scott and Martin Miller are also in the cast.\\ 
1. The screenwriter of Mix Me a Person was Ian Dalrymple. The drama genre of the film is Drama. The cast includes Carole Ann Ford, Donald Sinden, Tony Booth and Sergei Nolbandov.\\ 
2. The romantic comedy Dear Heart stars Richard Deacon, Barbara Nichols, Ruth McDevitt and Martin Manulis. It was written by Tad Mosel.\\ 
3. The Spy Who Dumped Me was produced by Brian Grazer and Lionsgate. It is rated B15 by the RTC. It stars Justin Theroux and Mila Kunis. It is written in English.\\ 
4. Lyra Belacqua is a fictional human being created by Philip Pullman. She is a film character portrayed by Dakota Blue Richards and Dafne Keen.\\ 
5. Waiting for the Light is a comedy film starring Teri Garr, Shirley MacLaine and produced by Caldecot Chubb. It was written by Christopher Monger.\\ 
6. Paper Heart is a comedy-drama starring Michael Cera and Demetri Martin. The film was written by Charlyne Yi and produced by Charlyne Yi. It also stars Derek Waters.\\ 
7. Philippa Boyens is the wife of Paul Gittins, who has a child called Calum Gittins.\\ 
8. I, Cesar is a drama film released on 9 April 2003. Its stars are Maria de Medeiros and Karine Silla. It was directed by Richard Berry.\\ 
9. Narc is a drama (film and television) with a mystery genre. Ray Liotta and Alan van Sprang star. Joe Carnahan is the screenwriter.\\ 
10. The Halfway House is a drama produced by Ealing Studios and distributed by Ealing Studios. It stars Glynis Johns and Françoise Rosay. The producer is Michael Balcon.\\ 
11. The Object of My Affection is a romantic comedy directed by Nicholas Hytner. It stars Peter Maloney and Sarah Hyland. The screenplay was written by Wendy Wasserstein.\\ 
12. Mr. Denning Drives North is a mystery film made by London Films. Its screenwriter is Alec Coppel, it stars John Mills and Phyllis Calvert. It was produced by Stephen Mitchell.\\ 
13. The screenwriter of The Girl and the Millionaire is Peer Guldbrandsen. Paul Hagen\ 
14. I Thank a Fool is a drama and mystery film written by Karl Tunberg. The cast includes Richard Wattis, Diane Cilento and Cyril Cusack.\\ 
15. The Wrong Man is a film noir written by Angus MacPhail. It stars Dayton Lummis, Esther Minciotti and Harold J. Stone.\\ 
16. Margot Kidder is married to Brian De Palma. \\ 
17. David Krumholtz starred in the episode "Scorched" in Numbers. He is the actor Charlie Eppes.\\ 
18. Amy Spettigue is a character in Where's Charley?.\\ 
19. The Impatient Alchemist is a 2002 mystery film directed by Patr cia Ferreira. Chete Lera and Miguel  ngel Sola are stars.\\ 
20. Martha O'Driscoll and Lou Costello are both stars, as is production designer John B. Goodman. Here Come the Co-Eds was also a TV series.\\ 
21. Mary Lynn Rajskub, Brent Spiner and Broderick Johnson are the stars of Dude, Where's My Car?\\ 
22. Philip Merivale is married to Gladys Cooper and is the husband of Viva Birkett. He has a child called John Merivale. Merivale is a human being.\\ 
23. The Wicker Man is a drama with a mystery genre. It stars Ross Campbell, Ingrid Pitt and Lindsay Kemp. The music is by Paul Giovanni.\\ 
24. The US film Catch 44 was produced by Megan Ellison. Its cast includes Malin  kerman, Deborah Ann Woll and Michael Benaroya. The film is about a Las Vegas Valley.\\ 

 \\
        \vspace{-1mm}
        \textbf{A:} Q-16; \\
        \bottomrule
    \end{tabular}
        \caption{
    Few-shot exemplars 5.
    }
    \label{tab:appendix-aqua-prompt}
\end{table}
\endgroup

\subsection{Evaluation Details}

We follow the previous work on the retrieval task and explanation graph task \cite{ZhilinYang2018HotpotQAAD,BhavanaDalvi2021ExplainingAW}, and we consider three evaluation indicators. They are the accuracy of interpretation graph construction, the recall of evidence retrieval, and the accuracy of NLDB QA.

Since the explanation graph can be expressed in many different forms, it needs to be evaluated comprehensively. Graph Matching (GM) is used to evaluate whether the structures of the two graphs are consistent. If all edges of two graphs are the same, they are considered to be consistent. We use the Graph Structure (GS) to evaluate whether the two graphs are isomorphic, which means that the two graphs may have different nodes but have the same graph structure. Graph Editing Distance \cite{ZeinaAbuAisheh2015AnEG} is a step of converting to a correct graph by calculating the addition, deletion, and replacement of nodes and edges, which can interpretably measure the distance between the predicted graph and the correct graph. Retrieval ability is essential to the task of NLDB. We use the F1 to valuate the retrieved evidence text. 

In recent years, more and more researchers use the language model to evaluate text \cite{TianyiZhang2019BERTScoreET,yuan2021bartscore}. It can evaluate the semantic level and has certain robustness. Therefore, we use the T5 model \cite{raffel2019exploring} \footnote{\url{https://huggingface.co/t5-base}} trained in the ReasonGraphQA answer generation task as a reader to deeply consider the quality of the retrieved evidence. Specifically, we provide the correct evidence and questions in the training set to the reader for training and then provide the retrieval results obtained by the retrieval system to the reader model for a generation. We use different random numbers to train three groups of readers, and take the complete matching rate of their generated results as the evaluation index to measure the accuracy of QA.


\end{document}